\PassOptionsToPackage{table}{xcolor}
\documentclass{streameval}
\usepackage{amsmath,amssymb}
\usepackage{tabularx,makecell,colortbl}
\usepackage{float,needspace,xspace}
\usepackage{tikz}
\newcommand*\circled[1]{\tikz[baseline=(char.base)]{
  \node[shape=circle,fill,font=\footnotesize,inner sep=0.4pt] (char) {\textcolor{white}{#1}};}}
\newcolumntype{Y}{>{\raggedright\arraybackslash}X}
\renewcommand{\arraystretch}{1.1}
\DeclareTextFontCommand{\textbf}{\bfseries}
\graphicspath{{figures/assets/}}
\newcommand{\ours}{StreamEval\xspace}
\newcommand{\papertitlefirst}{StreamEval: Evaluating Interactive}
\newcommand{\papertitlesecond}{Streaming Video Generation Models}
\newcommand{\papertitle}{\papertitlefirst\space\papertitlesecond}
\hypersetup{
  pdftitle={\papertitle},
  pdfauthor={StreamEval Team},
  pdfsubject={Interactive streaming video generation model evaluation},pdfkeywords={StreamEval, streaming video, benchmark}
}
\renewcommand{\titlefont}{\fontsize{20}{23}\selectfont}
\ifPDFTeX
  \newcommand{\coverresourcefont}{\fontfamily{qhv}\selectfont}
  \newcommand{\coverurlfont}{\fontfamily{zi4}\selectfont}
\else
  \newfontfamily\coverresourcefont{texgyreheros-regular.otf}[BoldFont=texgyreheros-bold.otf]
  \newfontfamily\coverurlfont{Inconsolatazi4-Regular.otf}
\fi
\renewcommand{\checkdatafont}{\fontsize{9}{11}\selectfont}
\renewcommand\checkdataformat[2][]{{\normalsize\coverresourcefont {\checkdatafont\bfseries #1:} #2}}
\definecolor{coverblue}{HTML}{286F98}
\definecolor{coverbackground}{HTML}{F0F5F8}
\renewcommand{\mymaketitle}{%
  \begingroup
  \renewcommand{\UrlFont}{\coverurlfont\mdseries\upshape}%
  \setlength{\parindent}{0pt}%
  \setlength{\parskip}{0pt}%
  \normalfont
  \vspace*{-13mm}%
\noindent
\begin{minipage}[c][10mm][c]{0.15\linewidth}\centering
  \includegraphics[width=\linewidth,height=9mm,keepaspectratio]{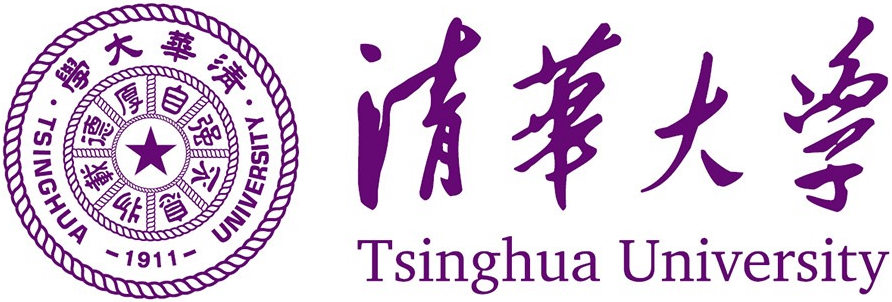}
\end{minipage}\hfill
\begin{minipage}[c][10mm][c]{0.1\linewidth}\centering
  \includegraphics[width=\linewidth,height=8mm,keepaspectratio]{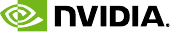}
\end{minipage}\hfill
\begin{minipage}[c][10mm][c]{0.14\linewidth}\centering
  \resizebox{\linewidth}{!}{%
    \raisebox{-0.5\height}{\includegraphics[height=10mm]{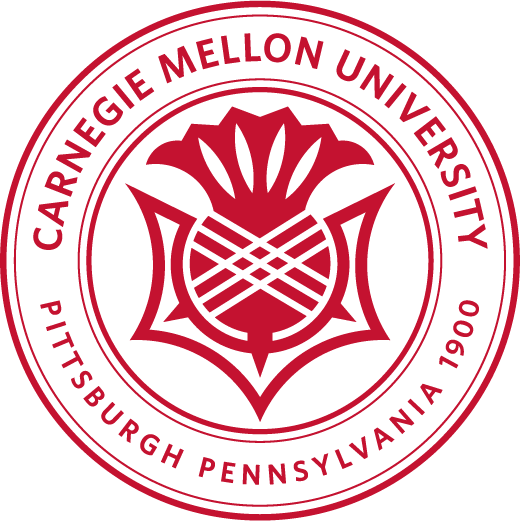}}%
    \hspace{1.5mm}%
    \raisebox{-0.5\height}{\includegraphics[height=9mm]{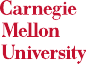}}%
  }
\end{minipage}\hfill
\begin{minipage}[c][10mm][c]{0.15\linewidth}\centering
  \resizebox{\linewidth}{!}{%
    \raisebox{-0.5\height}{\includegraphics[height=10mm]{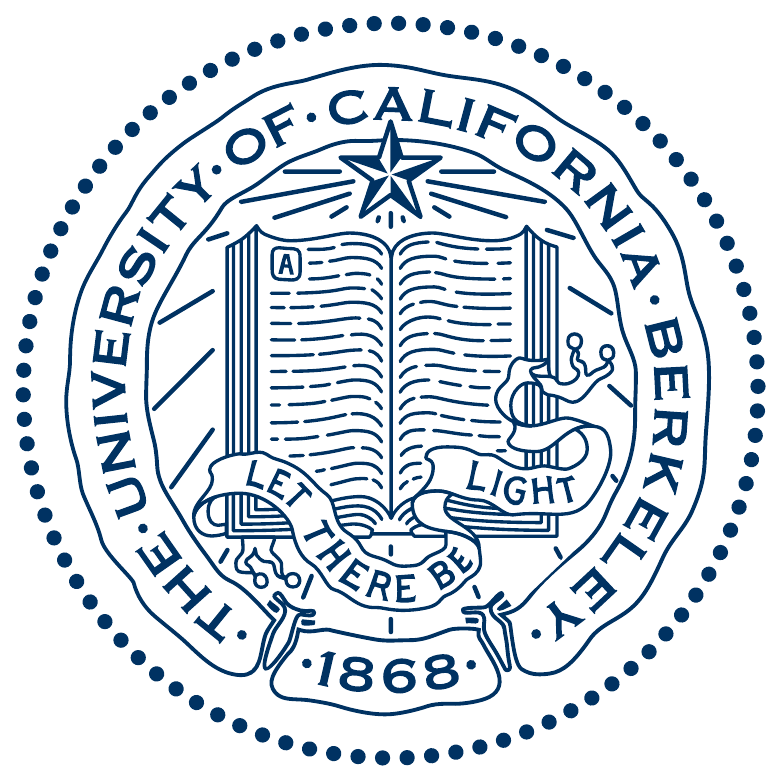}}%
    \hspace{1.5mm}%
    \raisebox{-0.5\height}{\includegraphics[width=18mm]{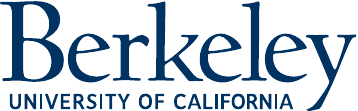}}%
  }
\end{minipage}\hfill
\begin{minipage}[c][10mm][c]{0.14\linewidth}\centering
  \includegraphics[width=\linewidth,height=10mm,keepaspectratio]{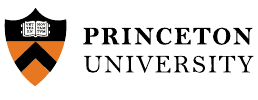}
\end{minipage}\hfill
\begin{minipage}[c][10mm][c]{0.115\linewidth}\centering
  \includegraphics[width=\linewidth,height=10mm,keepaspectratio]{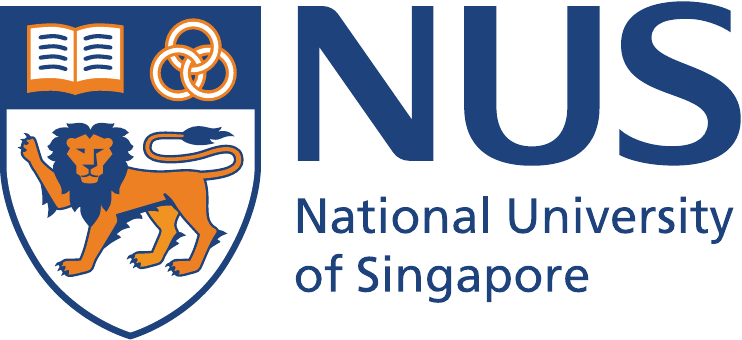}
\end{minipage}\hfill
\begin{minipage}[c][10mm][c]{0.14\linewidth}\centering
  \includegraphics[width=\linewidth,height=8mm,keepaspectratio]{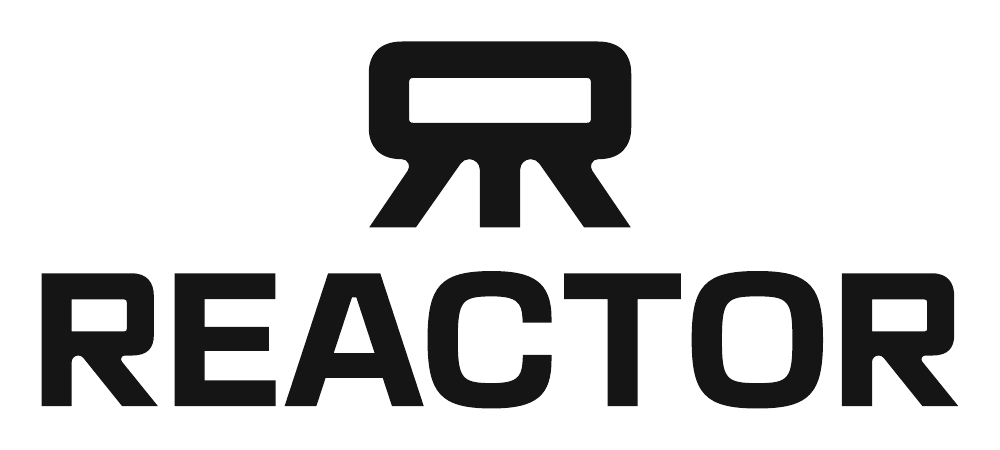}
\end{minipage}\par
  \vskip 8pt
  {\color{coverblue}\hrule height 1.3pt}%
  \vskip 8pt
  {\centering\titlelist\par}%
  \vskip 9pt
  {\color{coverblue}\hrule height 0.65pt}%
  \ifdefempty{\authorlist}{}{%
    \vskip 9pt
    {\centering\authorlist\par}}%
  \ifdefempty{\affiliationlist}{}{%
    \vskip 4pt
    {\centering\affiliationlist\par}}%
  \ifdefempty{\contributionlist}{}{%
    \vskip 4pt
    {\centering\contributionlist\par}}%
  \ifdefempty{\checkdatalist}{}{%
    \vskip 5pt
    {\centering\hypersetup{urlcolor=coverblue}\checkdatalist\par}}%
  \ifdefempty{\abstractlist}{}{%
    \vskip 9pt
    \begin{tcolorbox}[
      enhanced, colback=coverbackground, frame hidden,
      arc=2.5mm, boxrule=0pt, boxsep=0pt,
      left=5mm, right=5mm, top=3.5mm, bottom=4mm,
      before skip=0pt, after skip=0pt
    ]
      {\centering\abstractfont\sffamily\bfseries
        \color{black}Abstract\par}%
      \vskip 4pt
      \abstractlist\par
    \end{tcolorbox}}%
  \endgroup
  \thispagestyle{firststyle}%
  \par\vspace{8pt}%
}
\renewcommand{\maketitle}{%
  \begingroup
  \renewcommand{\UrlFont}{\coverurlfont\mdseries\upshape}%
  \setlength{\parindent}{0pt}%
  \setlength{\parskip}{0pt}%
  \normalfont
  \vspace*{-13mm}%
\noindent
\begin{minipage}[c][10mm][c]{0.15\linewidth}\centering
  \includegraphics[width=\linewidth,height=9mm,keepaspectratio]{figures/assets/affiliations/tsinghua-hd.png}
\end{minipage}\hfill
\begin{minipage}[c][10mm][c]{0.1\linewidth}\centering
  \includegraphics[width=\linewidth,height=8mm,keepaspectratio]{figures/assets/affiliations/nvidia.pdf}
\end{minipage}\hfill
\begin{minipage}[c][10mm][c]{0.14\linewidth}\centering
  \resizebox{\linewidth}{!}{%
    \raisebox{-0.5\height}{\includegraphics[height=10mm]{figures/assets/affiliations/cmu-symbol.png}}%
    \hspace{1.5mm}%
    \raisebox{-0.5\height}{\includegraphics[height=9mm]{figures/assets/affiliations/cmu-hd.pdf}}%
  }
\end{minipage}\hfill
\begin{minipage}[c][10mm][c]{0.15\linewidth}\centering
  \resizebox{\linewidth}{!}{%
    \raisebox{-0.5\height}{\includegraphics[height=10mm]{figures/assets/affiliations/berkeley-symbol.pdf}}%
    \hspace{1.5mm}%
    \raisebox{-0.5\height}{\includegraphics[width=18mm]{figures/assets/affiliations/berkeley.pdf}}%
  }
\end{minipage}\hfill
\begin{minipage}[c][10mm][c]{0.14\linewidth}\centering
  \includegraphics[width=\linewidth,height=10mm,keepaspectratio]{figures/assets/affiliations/princeton-hd.pdf}
\end{minipage}\hfill
\begin{minipage}[c][10mm][c]{0.115\linewidth}\centering
  \includegraphics[width=\linewidth,height=10mm,keepaspectratio]{figures/assets/affiliations/nus-hd.pdf}
\end{minipage}\hfill
\begin{minipage}[c][10mm][c]{0.14\linewidth}\centering
  \includegraphics[width=\linewidth,height=8mm,keepaspectratio]{figures/assets/affiliations/reactor.pdf}
\end{minipage}\par
  \vskip 8pt
  {\color{coverblue}\hrule height 1.3pt}%
  \vskip 8pt
  {\centering\titlelist\par}%
  \vskip 9pt
  {\color{coverblue}\hrule height 0.65pt}%
  \ifdefempty{\authorlist}{}{%
    \vskip 9pt
    {\centering\authorlist\par}}%
  \ifdefempty{\affiliationlist}{}{%
    \vskip 4pt
    {\centering\affiliationlist\par}}%
  \ifdefempty{\contributionlist}{}{%
    \vskip 4pt
    {\centering\contributionlist\par}}%
  \ifdefempty{\checkdatalist}{}{%
    \vskip 5pt
    {\centering\hypersetup{urlcolor=coverblue}\checkdatalist\par}}%
  \ifdefempty{\abstractlist}{}{%
    \vskip 9pt
    \begin{tcolorbox}[
      enhanced, colback=coverbackground, frame hidden,
      arc=2.5mm, boxrule=0pt, boxsep=0pt,
      left=5mm, right=5mm, top=3.5mm, bottom=4mm,
      before skip=0pt, after skip=0pt
    ]
      {\centering\abstractfont\sffamily\bfseries
        \color{black}Abstract\par}%
      \vskip 4pt
      \abstractlist\par
    \end{tcolorbox}}%
  \endgroup
  \thispagestyle{firststyle}%
  \par\vspace{8pt}%
}
\ifPDFTeX
\else
  \AtBeginDocument{}
\fi

\apptocmd{\thebibliography}{\interlinepenalty=10000\widowpenalty=10000\clubpenalty=10000}
  {}{\PackageError{streameval}{Unable to protect bibliography entries}{Check the bibliography environment.}}

\title{\papertitlefirst\\\papertitlesecond}
\date{}
\author[1,2,3,4,5,6,7]{StreamEval Team}
\affiliation[1]{Tsinghua University}
\affiliation[2]{NVIDIA}
\affiliation[3]{CMU}
\affiliation[4]{UC Berkeley}
\affiliation[5]{Princeton}
\affiliation[6]{NUS}
\affiliation[7]{Reactor}
\checkdata[Leaderboard]{\url{https://streameval.ai}}
\checkdata[Github Code]{\url{https://github.com/StreamEval/StreamEval}}
\checkdata[Benchmark Data]{\url{https://huggingface.co/datasets/StreamEval/StreamEval}}
\abstract{
\setlength{\parfillskip}{0pt}
We introduce \textbf{\ours} with four key features:
\circled{1}~\textbf{First interactive streaming video generation benchmark.}
\circled{2}~\textbf{Fully open source.} Benchmark data, evaluation code, protocols, and scoring rules are publicly available.
\circled{3}~\textbf{Comprehensive coverage.} Five tasks cover avatar dialogue, avatar performance, world generation, instruction editing, and reference editing, with 300 scripted cases lasting 15 to 300 seconds. Ten metrics assess task execution, state consistency, visual fidelity, and audiovisual coordination.
\circled{4}~\textbf{Monthly updates.} We will evaluate new and updated models and publish leaderboard results monthly.
Results from 38 models show that high visual quality and identity retention do not necessarily coincide with reliable request execution or sustained state consistency. The leaderboard is available at {\textbf{\url{https://streameval.ai}}}.\par
}

\begin{document}
\maketitle
\begingroup
\setlength{\intextsep}{0pt}
% First discussed in Section 1; fixed on the title page.
\begin{figure}[H]
  \centering
  \includegraphics[width=0.92\linewidth]{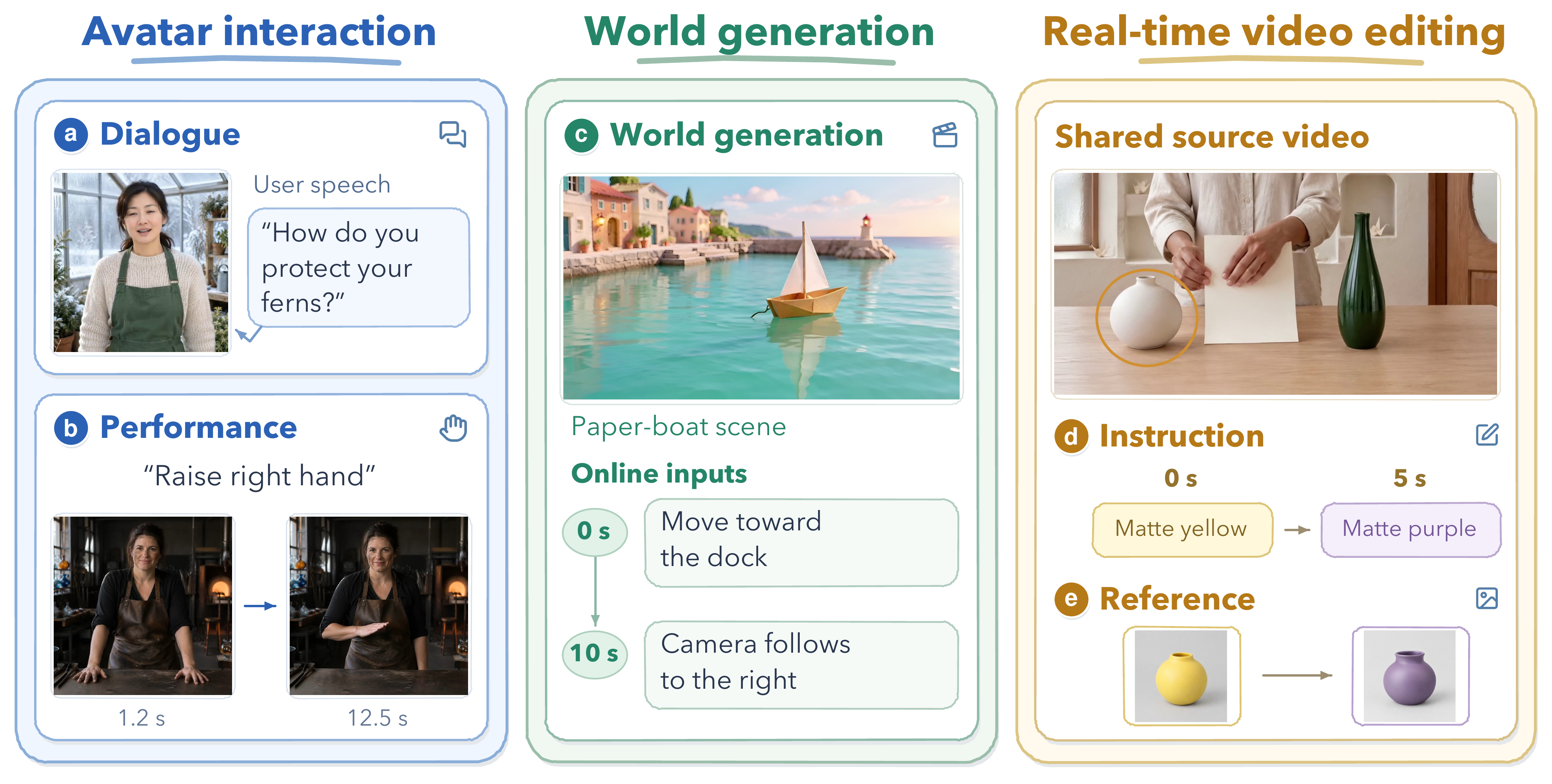}
  \caption{\textbf{StreamEval overview.} Five tasks span avatar interaction, world generation, and real-time video editing. Avatar panels show recorded frames; the world-generation panel illustrates a scenario. Editing uses a shared source video with text instructions or reference images that change over time.}
  \label{fig:overview}
\end{figure}

\endgroup

\clearpage
\tableofcontents
\clearpage

\section{Introduction}
\label{sec:introduction}

As streaming video generation advances, enabling real-time interaction with generated video is becoming increasingly important~\citep{zhang2026vidus2,xiao2026joyaivideoedit}. Systems that generate video incrementally can accept new user inputs while a session is in progress~\citep{yang2025longlive,chern2025livetalk}, supporting conversational avatars, controllable virtual worlds, and live video editing. Such settings require more than plausible individual frames or clips: generation and responsive control must remain aligned with incoming inputs as user requests and the surrounding context evolve over time.

Sustained interaction requires systems to execute new requests while preserving relevant prior content and state. A conversational avatar must respond in context without losing its identity or dialogue history; a generated world must accommodate new controls without disrupting unaffected scene content; and a video editor must apply, update, or remove edits while tracking an advancing source stream. Each request carries timing and scope requirements, defining when a change should occur, what it should modify, and what must remain intact. Evaluation must therefore assess both task execution and state consistency throughout a session alongside visual fidelity and audiovisual coordination.

Existing benchmarks examine complementary aspects of these requirements. VBench and EvalCrafter assess visual quality and temporal behavior, while T2V-CompBench examines compositional alignment~\citep{huang2024vbench,liu2024evalcrafter,sun2024t2vcompbench}. WBench and WBench-Editing evaluate multi-turn control, successive edits, and content preservation within generated worlds~\citep{ying2026wbench,liao2026editworld}. StreamAV-Bench assesses interactive updates and state retention in streaming video generation~\citep{liu2026streamavbench}, and LongV2VBench examines editing quality over sustained source streams~\citep{xiao2026joyaivideoedit}. VideoFDB evaluates full-duplex conversational agents through their perception and generation of nonverbal audiovisual cues~\citep{mazumdar2026videofdb}. Its focus on conversational dynamics complements task-directed evaluation of request execution and state preservation. Building on these efforts, we bring avatar interaction, world generation, and real-time video editing into a unified benchmark with explicit, task-specific execution and preservation criteria.

We introduce \textbf{\ours}, spanning five tasks across these domains: digital-human dialogue, avatar performance, world generation, instruction editing, and reference editing (\cref{fig:overview}). The benchmark contains 300 scripted test cases, each representing a 15--300\,s interactive session. A test case bundles the initial portrait, scene, or source video with a sequence of timed user requests and explicit criteria for what the output should change and preserve. For example, an editing case may change one object, update another while retaining the first edit, then remove an edit to restore the source. Inputs arrive incrementally, so the system must respond as the session unfolds. Ten task-specific metrics evaluate performance along four complementary dimensions: task execution, state consistency, visual fidelity, and audiovisual coordination. This framework structures evaluation across tasks while respecting differences in model input interfaces and metric applicability.

We evaluate 38 models on 300 test cases across five interactive streaming video tasks. Vidu S2 Avatar obtains the highest reported Dialogue score in digital-human dialogue. By Overall score, Vidu S2 Avatar leads avatar performance, MemFlow leads world generation, and Vidu S2 Editing leads both instruction and reference editing. Individual metrics reveal complementary strengths: LiveTalk obtains the highest Quality score in avatar performance, while JoyAI-Video-Edit obtains the highest Preservation score in instruction editing (\cref{sec:results}). These results highlight distinct model strengths across tasks and interaction requirements. Looking ahead, we view real-time interactive video as an important challenge for bringing omni-modal models to diverse application settings.

\enlargethispage{2\baselineskip}
Our contributions are:
\begin{itemize}
  \setlength{\itemsep}{2pt}
  \setlength{\parskip}{0pt}
  \item \textbf{Unified interactive benchmark.} Five streaming tasks and 300 scripted test cases specify timed multimodal inputs alongside explicit request-execution and state-preservation requirements.
  \item \textbf{Multidimensional evaluation.} Ten task-specific metrics assess task execution and state consistency alongside visual fidelity and audiovisual coordination under a shared session protocol.
  \item \textbf{Cross-task system study.} Evaluating 38 models across 300 test cases reveals limited instruction following in avatar performance and world generation, alongside low full-edit completion in reference editing.
\end{itemize}
\clearpage
\section{The StreamEval Benchmark}
\label{sec:benchmark}

\subsection{Motivation}
\label{sec:benchmark_motivation}

Real-time interactive video is an important step toward omni-modal systems that continuously perceive, generate, and respond to evolving user needs. Evaluating this capability requires testing whether systems execute new requests while preserving relevant context and state. \ours addresses these requirements through five tasks spanning avatar interaction, world generation, and real-time video editing, with task-specific metrics defined in \cref{sec:metrics}.

Each test case specifies initialization, a timeline of incoming inputs, and expected outcomes. A scheduler releases inputs incrementally and advances editing sources on the session clock. Input events specify when controls arrive; scoring opportunities specify what the system must accomplish or retain.

\paragraph{Avatar interaction.}
Digital-human dialogue requires contextually appropriate responses to user speech while maintaining persona and conversational history. Avatar performance requires rendering prescribed speech and requested visual actions, such as gaze shifts or hand gestures.

\paragraph{World generation.}
World generation creates evolving scenes in response to successive controls over objects, actions, cameras, or spatial state, while preserving unaffected content and continuity.

\paragraph{Real-time video editing.}
Instruction editing applies text-specified changes to an advancing video; reference editing uses images to specify the desired appearance. Both require modifying the intended target while preserving non-target content and source motion. Removal requests restore the contemporaneous source state.

\subsection{Benchmark Construction}
\label{sec:construction}

\begin{figure}[!htb]
  \centering
  \includegraphics[width=\linewidth]{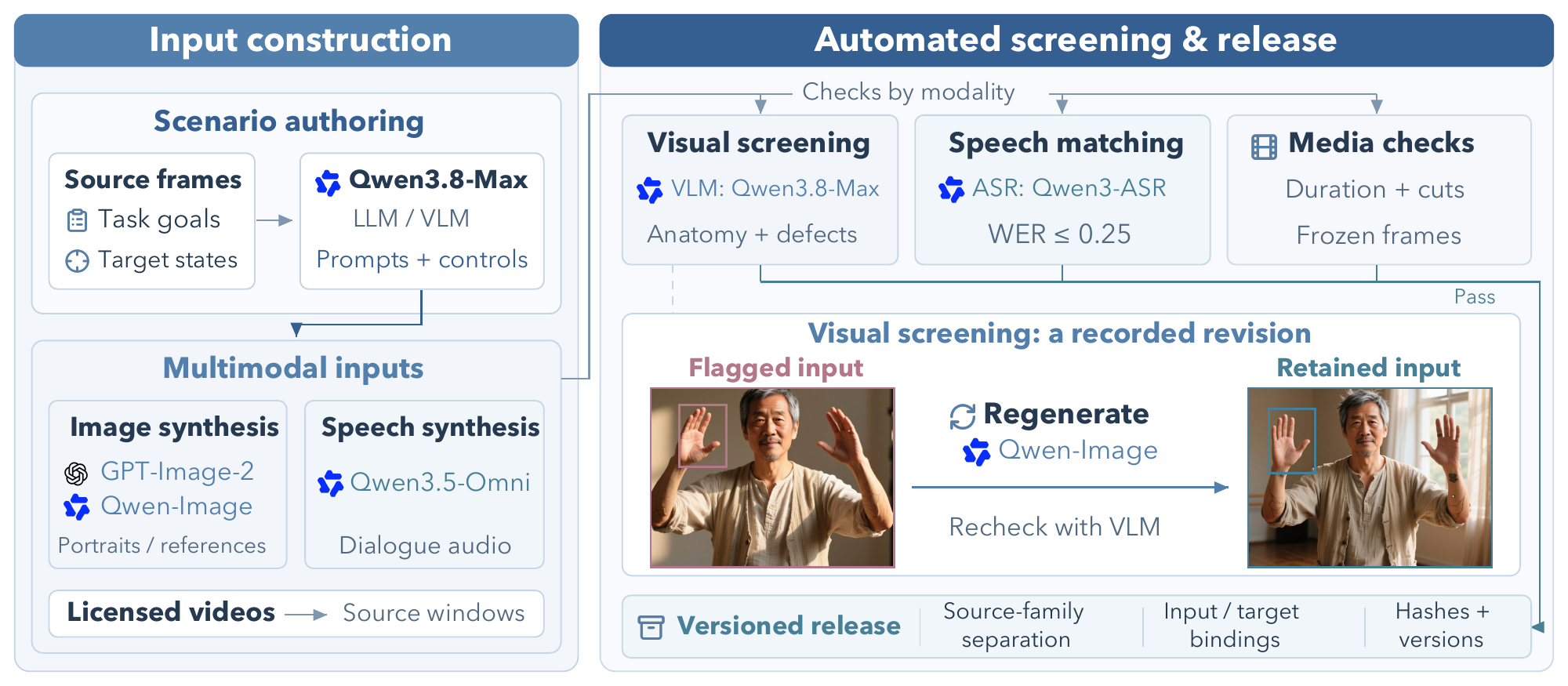}
  \caption{\textbf{Benchmark construction and verification.} Model-assisted authoring and input synthesis precede modality-specific checks. A Qwen-Image portrait with hand defects is regenerated and passes the same screening.}
  \label{fig:construction}
\end{figure}

We pair online inputs with explicit state requirements (\cref{fig:construction}). Qwen3.8-Max drafts session goals, timed controls, protected attributes, and expected outcomes. An authored oracle specifies what to establish, retain, cancel, or restore at each checkpoint. Media inputs are screened and revised against these requirements. All benchmark cases, including their timelines, oracles, and media, will be released together.

\paragraph{Functional diversity and difficulty.}
Each task covers individual operations, combined requests, and recovery. Dialogue covers factual answers, cross-turn state, turn-taking, and interruption; performance covers speech, gaze, expression, pose, and gestures. World generation tests object and action control, state updates, and camera or spatial continuity. Editing tests target establishment, incremental updates, reference switching, scope preservation, and restoration. Repair, multi-target updates, and reappearance after occlusion add more demanding interactions within the same benchmark.

Successive editing controls test whether established changes persist when new requests arrive. Cases include introducing a second target while retaining the first, updating selected attributes, switching references, and removing an edit to restore the advancing source.

\paragraph{Avatar and world-generation inputs.}
GPT-Image-2 and Qwen-Image synthesize portraits and visual references; portraits depict original synthetic adults or stylized fictional characters. Qwen3.5-Omni generates expressive user and prescribed speech. Dialogue pairs these recordings with authored personas and verified facts, including recordings with background noise or reverberation. Paired text-dialogue derivatives replace audio turns with authored text while retaining identical source-family identities.

\paragraph{Live editing sources.}
Editing tracks use continuous source videos comprising licensed real-world footage, animation, and benchmark-generated scenes. Third-party sources include Wikimedia Commons, NASA, and Blender Studio; per-asset licenses and attribution are supplied in the data release. Each source covers the session without looping, speed changes, or concatenation, so edits follow the advancing scene. Target regions are temporally grounded via Qwen3.8-Max at planned control timestamps.

\subsection{Data Statistics}
\label{sec:data_statistics}

\paragraph{Release composition.}
Each task contains 60 test cases, balanced across 15, 30, 60, 120, and 300 seconds with 12 cases per duration. Across all five tracks, 300 test cases span 240 independent source families, with 60 text-dialogue derivatives reusing existing families. One pass through the 300 video-session cases represents 525 minutes of active streaming generation, excluding setup and response tails.

\begin{figure}[tb]
  \centering
  \includegraphics[width=\linewidth]{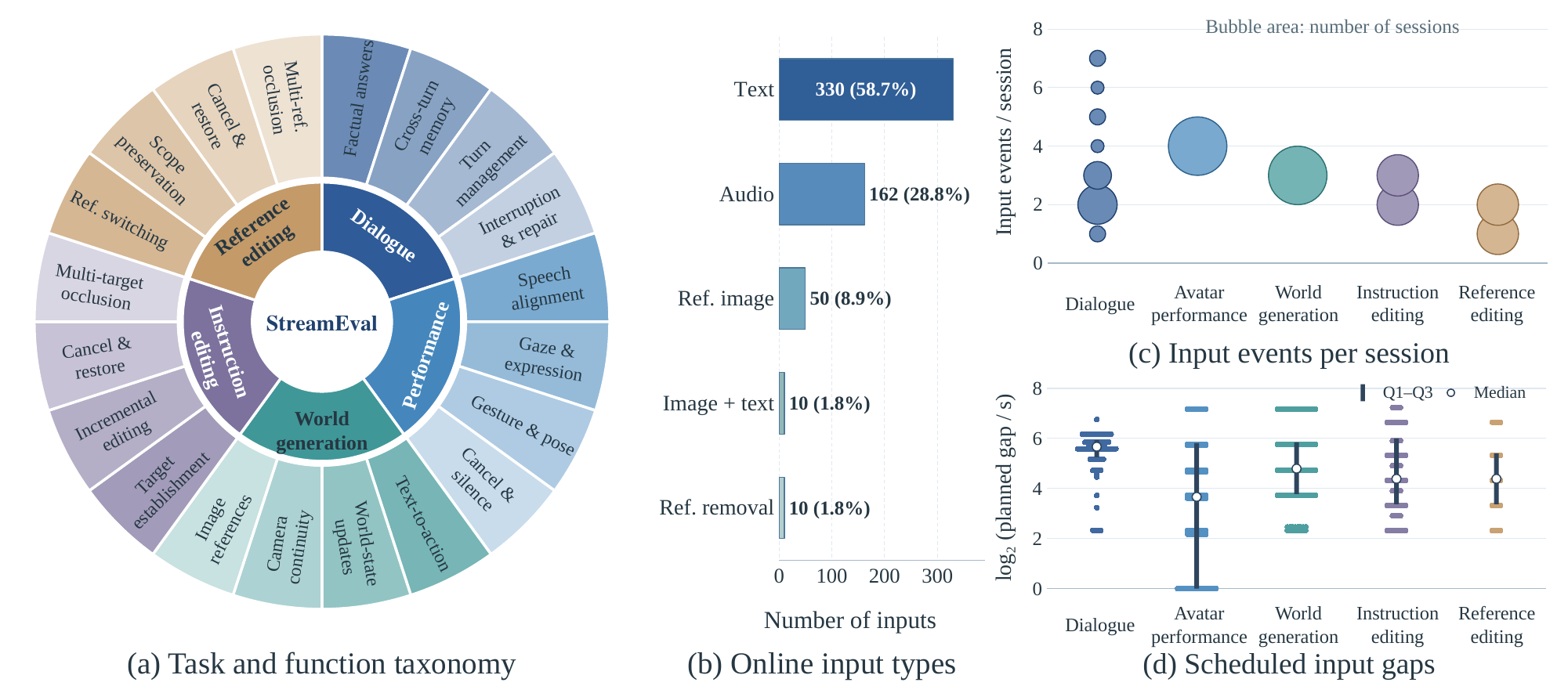}
  \caption{\textbf{Benchmark composition.} (a) Task and function composition. (b) Planned input types. (c) Inputs per session. (d) Gaps between adjacent scheduled inputs. The 200 cases shown contain 562 online events, excluding initialization and continuous source frames. Gap statistics exclude \mbox{pairs containing a trigger.}}
  \label{fig:statistics}
\end{figure}

\paragraph{Interaction load and diversity.}
The 300 test cases contain 1,347 planned online events, excluding initialization and continuous source frames. Inputs include speech, text, and visual references; each input can impose multiple execution and preservation requirements. Avatar designs span photorealistic, stylized 3D, anime, and clay or stop-motion appearances. Dialogue includes clean speech, background noise, and room reverberation, while editing covers indoor and outdoor scenes with single or multiple targets. \Cref{fig:statistics} summarizes interaction statistics for 200 of these cases.

\begin{table}[htb]
\centering
\caption{\textbf{Five-task coverage and evaluation duration.} Dlg.: dialogue; Perf.: avatar performance; World: world generation; Instr./Ref.: instruction/reference real-time video editing. \ensuremath{\checkmark}/\ensuremath{\times} denote task inclusion/exclusion.}
\label{tab:benchmark_comparison}
\definecolor{ScopeBlue}{HTML}{365F91}
\definecolor{ScopeGray}{HTML}{7D8590}
\newcommand{\scopeYes}{\textcolor{ScopeBlue}{\ensuremath{\checkmark}}}
\newcommand{\scopeNo}{\textcolor{ScopeGray}{\ensuremath{\times}}}
\setlength{\tabcolsep}{7pt}
\renewcommand{\arraystretch}{1.2}
\begingroup
\small
\resizebox{0.7\linewidth}{!}{%
\begin{tabular}{lcccccc}
\toprule
\multirow{2}{*}[-0.35ex]{\textbf{Benchmark}}
& \multicolumn{5}{c}{\textbf{Task coverage}}
& \multirow{2}{*}[-0.35ex]{\textbf{Duration (s)}} \\
\cmidrule(lr){2-6}
& \textbf{Dlg.} & \textbf{Perf.} & \textbf{World} & \textbf{Instr.} & \textbf{Ref.}
& \\
\midrule
VBench~\citep{huang2024vbench}
& \scopeNo & \scopeNo & \scopeYes & \scopeNo & \scopeNo & 2--3.3 \\
VBench++~\citep{huang2024vbenchpp}
& \scopeNo & \scopeNo & \scopeYes & \scopeNo & \scopeNo & Var. \\
T2V-CompBench~\citep{sun2024t2vcompbench}
& \scopeNo & \scopeNo & \scopeYes & \scopeNo & \scopeNo & 2--5 \\
TC-Bench~\citep{feng2024tcbench}
& \scopeNo & \scopeNo & \scopeYes & \scopeNo & \scopeNo & Var. \\
NarraStream-Bench~\citep{liu2026iamflow}
& \scopeNo & \scopeNo & \scopeYes & \scopeNo & \scopeNo & 60 \\
MBench~\citep{zhang2026mbench}
& \scopeNo & \scopeNo & \scopeYes & \scopeNo & \scopeNo & 25 \\
WorldMark~\citep{xu2026worldmark}
& \scopeNo & \scopeNo & \scopeYes & \scopeNo & \scopeNo & 20--60 \\
WBench~\citep{ying2026wbench}
& \scopeNo & \scopeNo & \scopeYes & \scopeNo & \scopeNo & Var. \\
WBench-Editing~\citep{liao2026editworld}
& \scopeNo & \scopeNo & \scopeYes & \scopeNo & \scopeNo & 15--30 \\
StreamAV-Bench~\citep{liu2026streamavbench}
& \scopeNo & \scopeNo & \scopeYes & \scopeNo & \scopeNo & 180 \\
\addlinespace[4pt]
GenBench~\citep{huang2025liveavatar}
& \scopeNo & \scopeYes & \scopeNo & \scopeNo & \scopeNo & $\sim$10; $>300$ \\
THEval~\citep{quignon2025theval}
& \scopeNo & \scopeYes & \scopeNo & \scopeNo & \scopeNo & 2--334.6 \\
Full-Duplex-Bench~\citep{lin2025fullduplexbench}
& \scopeYes & \scopeNo & \scopeNo & \scopeNo & \scopeNo & Var. \\
VideoFDB~\citep{mazumdar2026videofdb}
& \scopeYes & \scopeNo & \scopeNo & \scopeNo & \scopeNo & Var. \\
\addlinespace[4pt]
EditVerseBench~\citep{ju2025editverse}
& \scopeNo & \scopeNo & \scopeNo & \scopeYes & \scopeYes & Var. \\
LiveEdit~\citep{wang2026liveedit}
& \scopeNo & \scopeNo & \scopeNo & \scopeYes & \scopeNo & -- \\
EgoEditBench~\citep{li2025egoedit}
& \scopeNo & \scopeNo & \scopeNo & \scopeYes & \scopeNo & Var. \\
CharEdit-Bench~\citep{li2026editalive}
& \scopeNo & \scopeNo & \scopeNo & \scopeYes & \scopeNo & 5; $>60$ \\
LongV2VBench~\citep{xiao2026joyaivideoedit}
& \scopeNo & \scopeNo & \scopeNo & \scopeYes & \scopeNo & 60 \\
\midrule
\rowcolor{ScopeBlue!8}
\textbf{\ours}
& \scopeYes & \scopeYes & \scopeYes & \scopeYes & \scopeYes
& \textbf{15--300} \\
\bottomrule
\end{tabular}%
}
\endgroup
\par\vspace{4pt}
\begin{minipage}{0.98\linewidth}
\fontsize{8}{9.6}\selectfont
Duration follows the benchmark evaluation, excluding longer model demonstrations; Var.: model/case-dependent; --: unspecified. Dialogue includes audio-only and audiovisual conversation. VideoFDB's avatar responses are conversational, not prescribed-performance tasks; its duration varies by clip. Online control follows each benchmark's input protocol.
\end{minipage}
\end{table}

\paragraph{Evaluation roster.}
The evaluation protocol specifies a roster of 300 test cases: 60 per task, with 12 at each duration tier and three independent runs per case and model. The results reported in this paper cover these 300 cases (\cref{sec:experimental_setup}). All five duration strata receive equal aggregation weight within each task, so longer sessions contribute equally to the final assessment rather than in proportion to their duration.

\subsection{Comparison with Existing Benchmarks}
\label{sec:benchmark_comparison}

\Cref{tab:benchmark_comparison} positions \ours against existing video generation and interactive benchmarks across task coverage and evaluation horizons. Avatar-performance benchmarks assess rendered speakers, including visual quality and speech-motion synchronization, while Full-Duplex-Bench evaluates spoken dialogue without video output. VideoFDB additionally evaluates full-duplex conversational perception and audiovisual response generation~\citep{mazumdar2026videofdb}. Its naturally occurring nonverbal cues complement our authored speech-input dialogue and prescribed-performance tasks; our broader task coverage does not imply coverage of its user-video perception requirements. For video editing, WBench and WBench-Editing focus on world-model environments, while EditVerseBench covers instruction- and reference-conditioned editing of external videos~\citep{ying2026wbench,liao2026editworld,ju2025editverse}. WBench-Editing also reports a separate reference-conditioned comparison. Task interfaces and control timing jointly define the interaction protocol.

\ours applies a balanced multi-horizon design across all five streaming tasks: 15, 30, 60, 120, and 300 seconds, with 60 test cases at each duration tier. Its 300-second editing cases represent a five-fold expansion over LongV2VBench (60\,s), while its world-generation cases extend beyond StreamAV-Bench's 180-second rollouts~\citep{xiao2026joyaivideoedit,liu2026streamavbench}. Longer avatar sequences already appear in GenBench~\citep{huang2025liveavatar}; our contribution is balanced coverage across tasks and durations, with incrementally released multimodal controls and continuously advancing editing sources.

We evaluate 38 models on 300 test cases across these tasks (\cref{sec:experimental_setup}). For context, WBench evaluates 20 models, WorldMark 10, and StreamAV-Bench 13 systems within their respective domains~\citep{ying2026wbench,xu2026worldmark,liu2026streamavbench}. Our benchmark brings these application domains into a shared framework for assessing request execution and state preservation alongside content quality.

\subsection{Quality Verification}
\label{sec:quality_verification}

\paragraph{Asset and consistency checks.}
Automated checks verify media decodability and duration bounds, file identity, and agreement among events, opportunities, and scoring requirements (\cref{fig:construction}). Qwen3-ASR transcribes synthesized speech, requiring normalized word-error rates of at most 0.25 against authored transcripts. Qwen3.8-Max audits images for subject clarity and anatomical or rendering defects. Editing sources are checked for scene cuts, frozen spans, and unsuitable targets. Revised candidates must pass the same checks.

\paragraph{Judge calibration.}
Targeted positive and negative probes evaluate automated production judges against acceptance criteria of at most 5\% false positive rate, 10\% false negative rate, and 10\% unresolved decisions per class. Visual critique judges return structured defect categories, severity ratings, and supporting frame timestamps. Identity uses the lower rating from two judges given the same initialization portrait (or the system's opening frame for a fixed avatar) and sampled output frames to guard against same-style identity substitutions.

\FloatBarrier

\FloatBarrier
\section{System Evaluation}
\label{sec:evaluation}
\suppressfloats[t]

We evaluate how systems execute evolving requests, retain relevant state, and render the resulting video across the five tasks in \cref{sec:benchmark}. Each task and input setting has a defined metric profile and an Overall score.

\subsection{Evaluation Setup}
\label{sec:experimental_setup}
\label{sec:tasks}

\paragraph{Evaluated models.}
We report 38 distinct models across 49 task configurations. A task configuration is a model evaluated on one task with its specified input route and any adapter components. A model evaluated on multiple tasks is counted once in the model total and separately for each task configuration. Avatar models include Vidu S2 Avatar~\citep{zhang2026vidus2}, Tavus CVI Full, LiveAvatar FULL, Runway Characters, Vivix A1, AptAvatar, Avatar-Forever, AvatarForcing, AVTR-1, LiveTalk~\citep{chern2025livetalk}, MuseTalk~\citep{zhang2025musetalk}, Quark Live Avatar, and the SoulX-FlashHead, SoulX-FlashTalk, and SoulX-LiveAct models. World generation covers AlayaWorld, HappyOyster 1.0 Adventure, HY-World 1.5 WorldPlay, LingBot-World 2.0, Matrix-Game 3.5, Causal Forcing, Context Forcing, Helios, Krea Realtime, LongLive 2.0, MemFlow, Reward Forcing, Rolling Forcing~\citep{liu2025rollingforcing}, Self-Forcing, ShotStream, Yume 1.5, and Zing-0.5. Reported editing models include Vidu S2 Editing, JoyAI-Video-Edit~\citep{xiao2026joyaivideoedit}, EditaLive, Lucy 2.5 (Decart), SANA-Streaming, and XMax X2.0. The editing tables show configurations with Overall at least 5; this presentation subset was selected after evaluation. The model and configuration totals refer to the reported subset. LiveAvatar FULL and Tavus CVI Full use fixed, provider-supplied avatars because avatar customization is unavailable through their evaluated interfaces; Identity is rated against each system's own opening frame and included in Overall, and their names appear in gray in \cref{tab:results_digital_human}.

\paragraph{Evaluation tasks.}
The five tasks are digital-human dialogue (15 configurations), avatar performance (7), world generation (17: 12 text-conditioned and 5 image-conditioned), instruction editing (6), and reference editing (4). The evaluation covers 300 test cases (60 in each task) across five equally weighted duration strata from 15 to 300\,s, with three repeats per test case and model. World generation includes text-controlled and image-conditioned inputs; the latter adds appearance conditioning.

\paragraph{Evaluation metrics.}
We report ten metrics across four dimensions: task execution (Dialogue, Following, Edit Success), state consistency (Stability, Continuity, Preservation), visual fidelity (Quality, Identity, Reference), and audiovisual coordination (A--V Sync). Each task reports the equal-weight mean of its applicable metrics as Quality-Overall; avatar tasks additionally report a customization bonus and their sum as Overall.

\subsection{Metrics and Scoring}
\label{sec:metrics}

\subsubsection{Task-Specific Metrics}

\paragraph{Shared output measures.}
All five tasks assess output quality and stability under the expected state sequence, including authorized actions, edits, and restoration. Quality scores five-second clips from primary-judge defect lists: no defect, one minor defect, multiple minor defects, one severe defect, and multiple severe defects give ratings $r \in \{5, 4, 3, 2, 1\}$, respectively, with severe defects taking precedence and mapping to $25(r-1)$. Stability measures the duration covered by clips with no severe temporal defect, at most one minor temporal defect, and Quality rating at least 3. Because Stability is gated by Quality, the two metrics overlap and a high Stability score requires both temporal consistency and acceptable visual quality.

\paragraph{Digital-human dialogue.}
Dialogue measures the percentage of planned turns satisfying answer, content, conversational, and timing requirements. Identity provides the initialization portrait (or the system's own opening frame for a fixed avatar) and sampled clip frames to two judges, Qwen3.8-Max-0902 and Qwen3-VL-Plus, and takes the lower feature-derived rating while allowing authorized expression, pose, crop, and lighting changes. An explicitly different person or at least three clear feature changes gives rating 1; two clear changes give 2; one clear change or at least three slight changes gives 3; one or two slight changes give 4; no change gives 5. An unclear same-person judgment caps the rating at 3, and the final score is $25(r-1)$. A--V Sync measures the percentage of required speech events with lip-opening/audio-envelope correlation at least 0.3, evaluated at a pooled system offset whose 95\% interval (2,000 session-bootstrap draws) lies within $[-200,200]$\,ms using 10-ms search steps over $[-1000,1000]$\,ms. Sparse source sampling (500\,ms for dialogue, 250\,ms for performance) limits its temporal resolution. Together with Quality and Stability, these metrics evaluate conversational execution, speaker appearance, and audiovisual delivery.

\paragraph{Avatar performance.}
Following measures the percentage of control events in which the avatar executes the requested gaze, expression, pose, or gesture on the specified target. Prescribed speech supplies the verbal content. The task combines Following with Quality, Stability, Identity, and A--V Sync.

\paragraph{World generation.}
Following measures the percentage of online controls that produce the requested scene operation on the specified target; a control passes when every frame in which the target is visible shows the requested outcome fully or partially. Continuity measures the fraction of generated duration meeting motion, camera, object-permanence, transition, and spatial-relation requirements. These two metrics join Quality and Stability in both the text-conditioned and the image-conditioned setting.

\paragraph{Instruction editing.}
Following checks the requested operation and target, allowing a partial appearance match when no severe leakage occurs. Edit Success requires the complete edit to satisfy scope, active-state, deadline, persistence, and protected-content checks. Preservation assesses non-target content and current source motion, including restoration against the contemporaneous source. Together with Quality and Stability, this profile separates operation execution, full edit completion, source retention, and visual fidelity.

\paragraph{Reference editing.}
Reference measures declared appearance attributes at checkpoints where the supplied image is active, whereas avatar Identity compares against the initialization portrait or, for fixed-avatar interfaces, the system's opening frame. Edit Success checks the full editing contract, Preservation assesses protected source content and motion, and Quality and Stability complete the five-metric profile.

\begin{figure}[!htb]
  \centering
  \includegraphics[width=\linewidth]{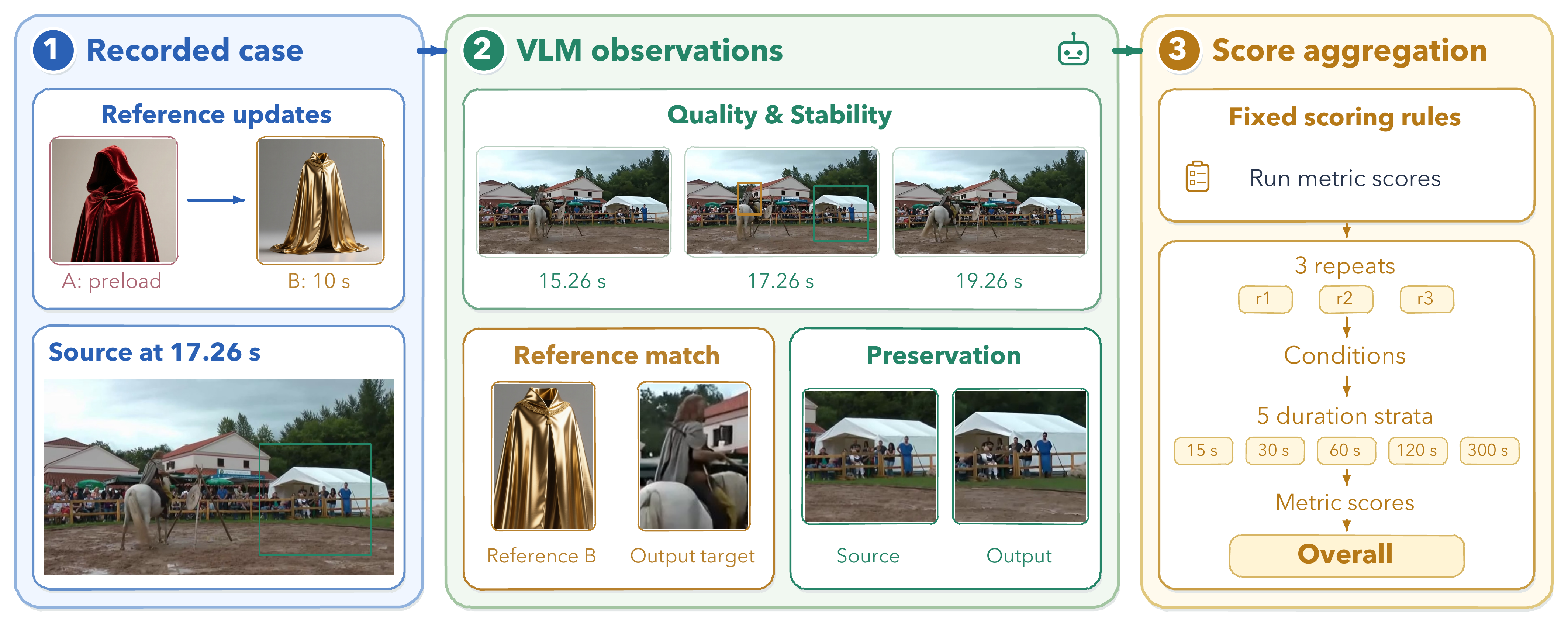}
  \caption{\textbf{Evaluation workflow.} Recorded inputs and output frames from reference-editing case 234 illustrate quality, stability, reference, and preservation checks. Fixed rules convert structured VLM observations into run metric scores. Aggregation averages repeats, conditions, and duration strata before computing \mbox{task-specific Overall.}}
  \label{fig:evaluation_framework}
\end{figure}

Fixed rules convert structured visual observations into scores using each test case's predefined deadline and verification window. \Cref{fig:evaluation_framework} illustrates a recorded reference-editing session.

\FloatBarrier
\Needspace{8\baselineskip}

\subsubsection{Score Aggregation}
\label{sec:aggregation_observation}

The ten content metrics use a 0--100 scale, with applicability fixed before inspecting outputs. Missing output and unsupported required operations receive zero, and all planned control opportunities remain in the denominator, including unfired triggers and controls beyond a provider's session limit. Unresolved judgments define an interval by contributing 0 to the lower endpoint and 100 to the upper endpoint; tables report the lower endpoint when any required judgment is unresolved. Let $s_{c,r,m}$ denote the run-level score for test case $c$, repeat $r$, and metric $m$, and let $\mathcal{C}_{d,m}$ contain the applicable test cases in duration stratum $d$. We average the three repeats, then test cases within each stratum, and finally the five duration strata $\mathcal{D}$:
\begin{equation}
S_m = \frac{1}{|\mathcal{D}|}\sum_{d\in\mathcal{D}}
\frac{1}{|\mathcal{C}_{d,m}|}\sum_{c\in\mathcal{C}_{d,m}}
\frac{1}{3}\sum_{r=1}^{3}s_{c,r,m}.
\label{eq:aggregation}
\end{equation}

\paragraph{Task-specific Overall.}
For task $t$ and model $k$, let $\mathcal{M}_{t,k}$ contain its applicable metrics and $S_{t,k,m}$ denote the metric aggregate from \cref{eq:aggregation}. Quality-Overall is their equal-weight mean:
\begin{equation}
\operatorname{QOverall}_{t,k}
=\frac{1}{|\mathcal{M}_{t,k}|}\sum_{m\in\mathcal{M}_{t,k}}S_{t,k,m}.
\label{eq:overall}
\end{equation}
We abbreviate Quality-Overall as Q-Overall. Portrait-conditioned dialogue, performance, and both editing tasks average five metrics; fixed-avatar dialogue also averages five, because Identity is rated against the system's own opening frame when its evaluated interface does not accept a custom portrait. World generation averages four metrics.

\paragraph{Sweet-spot bonus for avatar customization.}
Avatar customization is an essential capability for digital-human dialogue and avatar performance, so these two tasks add a sweet-spot bonus to Q-Overall. Every system that supports a custom avatar receives $+5$; systems restricted to fixed, provider-supplied avatars receive 0. The tables report Q-Overall, the bonus (Custom.), and their sum as Overall, which orders the avatar tasks. All other tasks report Overall $=$ Q-Overall. Thus,
\begin{equation}
\operatorname{Overall}_{t,k}=\operatorname{QOverall}_{t,k}+B_{t,k},\qquad
B_{t,k}=\begin{cases}
5, & \text{avatar task with custom-avatar support},\\
0, & \text{otherwise}.
\end{cases}
\end{equation}
Q-Overall remains on a 0--100 scale; the avatar ranking score can reach 105. The bonus describes interface capability and does not alter the underlying metric scores.

\subsection{Experimental Results}
\label{sec:results}

\Cref{tab:results_digital_human,tab:results_avatar_performance,tab:results_general_interaction,tab:results_instruction_edit,tab:results_reference_edit} report model performance across the five tasks in alphabetical order. The website shows the same scores sorted by descending Overall within each task.

\subsubsection{Digital-Human Dialogue}
\label{sec:results_dialogue}

Among custom-avatar models, Vidu S2 Avatar achieves the highest Overall score (93.98; 88.98 Q-Overall plus the customization bonus) and leads in Quality (96.78), Dialogue (75.29), and Identity (90.12), followed by AVTR-1 at 84.14 Overall (79.14 Q-Overall), 82.97 Quality, and 88.54 Identity (\cref{tab:results_digital_human}). In the fixed-avatar setting, where Identity is rated against each system's own opening frame, Tavus CVI Full reaches 86.65 Q-Overall and 89.89 Identity compared with 65.07 and 87.70 for LiveAvatar FULL; neither supports a custom avatar, so both receive no bonus and their Overall equals Q-Overall. Across individual shared dimensions, Tavus CVI Full records the highest A--V Sync (94.44 vs.\ 84.87 for Vidu S2 Avatar) and the second-highest Dialogue (73.91, behind Vidu S2 Avatar at 75.29 and ahead of 61.37 for LiveAvatar FULL), while LiveAvatar FULL reaches 100.00 in Stability alongside 0.00 in A--V Sync.

Results among models evaluated with the shared dialogue pipeline further show that strong visual appearance and temporal stability do not automatically accompany high synchronization or conversational scores. MuseTalk 1.5~\citep{zhang2025musetalk} attains 95.34 in Quality, 97.11 in Stability, and 88.44 in Identity, yet records 38.09 in A--V Sync and 54.34 in Dialogue (79.66 Overall). Conversely, AptAvatar 14B reaches 70.32 in A--V Sync and 94.74 in Stability, but scores 35.98 in Dialogue and 68.03 in Quality (75.01 Overall), whereas SoulX-FlashTalk 14B records 68.90 Quality, 66.27 A--V Sync, and 50.57 Dialogue to reach 77.90 Overall.

\begin{table}[htb]
\centering
\caption{\textbf{Digital-human dialogue.}}
\label{tab:results_digital_human}
\definecolor{tableAccent}{HTML}{365F91}
\fontsize{8.5}{10.2}\selectfont
\setlength{\tabcolsep}{2pt}
\renewcommand{\arraystretch}{1.18}
\begin{tabularx}{\linewidth}{@{}Yrrrrrrrr@{}}
\toprule
\textbf{Model} & \textbf{Quality} & \textbf{A--V Sync} & \textbf{Stability} & \textbf{Dialogue} & \textbf{Identity} & \textbf{Q-Overall} & \textbf{Custom.} & \textcolor{tableAccent}{\textbf{Overall}} \\
\midrule
AptAvatar 14B$^{\dagger}$ & 68.03 & 70.32 & 94.74 & 35.98 & 81.00 & 70.01 & +5 & 75.01 \\
Avatar-Forever 22B$^{\dagger}$ & 50.86 & 48.17 & 68.48 & 54.93 & 75.23 & 59.53 & +5 & 64.53 \\
AvatarForcing 1.3B$^{\dagger}$ & 57.17 & 0.00 & 87.30 & 8.93 & 66.85 & 44.05 & +5 & 49.05 \\
AVTR-1$^{\dagger}$ & 82.97 & 69.38 & 99.87 & 54.97 & 88.54 & 79.14 & +5 & 84.14 \\
\textcolor{black!55}{LiveAvatar FULL} & 76.27 & 0.00 & 100.00 & 61.37 & 87.70 & 65.07 & 0 & 65.07 \\
LiveTalk 1.3B$^{\dagger}$ & 61.75 & 62.80 & 86.51 & 54.40 & 77.51 & 68.60 & +5 & 73.60 \\
MuseTalk 1.5$^{\dagger}$ & 95.34 & 38.09 & 97.11 & 54.34 & 88.44 & 74.66 & +5 & 79.66 \\
Quark Live Avatar 14B$^{\dagger}$ & 56.00 & 64.91 & 70.90 & 55.02 & 78.10 & 64.99 & +5 & 69.99 \\
Runway Characters & 57.81 & 49.82 & 75.94 & 43.53 & 60.40 & 57.50 & +5 & 62.50 \\
SoulX-FlashHead (Lite)$^{\dagger}$ & 68.43 & 29.77 & 91.61 & 56.65 & 79.69 & 65.23 & +5 & 70.23 \\
SoulX-FlashTalk 14B$^{\dagger}$ & 68.90 & 66.27 & 95.45 & 50.57 & 83.31 & 72.90 & +5 & 77.90 \\
SoulX-LiveAct 18B$^{\dagger}$ & 64.93 & 50.30 & 91.54 & 54.98 & 75.09 & 67.37 & +5 & 72.37 \\
\textcolor{black!55}{Tavus CVI Full} & 75.04 & 94.44 & 99.95 & 73.91 & 89.89 & 86.65 & 0 & 86.65 \\
Vidu S2 Avatar & 96.78 & 84.87 & 97.83 & 75.29 & 90.12 & 88.98 & +5 & 93.98 \\
Vivix A1 & 45.78 & 69.15 & 66.83 & 50.38 & 75.19 & 61.47 & +5 & 66.47 \\
\bottomrule
\end{tabularx}
\par\vspace{2pt}{\scriptsize Q-Overall: equal-weight mean of the content metrics. Custom.: sweet-spot bonus, +5 for systems that support a custom avatar and 0 otherwise. Overall $=$ Q-Overall $+$ Custom.}\par\vspace{3pt}
\end{table}

\FloatBarrier
\Needspace{9\baselineskip}
\subsubsection{Avatar Performance}
\label{sec:results_performance}

Vidu S2 Avatar (81.71 Overall) and LiveTalk 1.3B (75.85 Overall) lead avatar performance with distinct metric profiles (\cref{tab:results_avatar_performance}). Vidu S2 Avatar achieves the highest scores in A--V Sync (99.60), Stability (97.25), Identity (86.12), and Following (36.90), paired with 63.68 in Quality. LiveTalk 1.3B leads in Quality (71.27) and reaches 16.59 in Following, while recording 86.44 in A--V Sync, 96.06 in Stability, and 83.89 in Identity. AptAvatar 14B ranks third at 70.37 Overall, ahead of Quark Live Avatar 14B at 62.83 and SoulX-LiveAct 18B at 53.51; all seven systems support a custom avatar and receive the $+5$ bonus.

Middle-ranking systems also differ in temporal stability and visual quality despite similar lip synchronization and identity preservation scores. AptAvatar 14B and Quark Live Avatar 14B tie at 98.89 in A--V Sync and score similarly in Identity (81.33 vs.\ 79.75), but differ in Stability (78.06 vs.\ 55.30) and Quality (55.55 vs.\ 43.51). Across all seven evaluated models, executing scripted actions under timing constraints remains the primary challenge: Following scores span 6.37 (Avatar-Forever 22B) to 36.90 (Vidu S2 Avatar), even for systems that exceed 80 in Identity and 96 in Stability.

\begin{table}[!htb]
\centering
\caption{\textbf{Avatar performance.}}
\label{tab:results_avatar_performance}
\definecolor{tableAccent}{HTML}{365F91}
\fontsize{8.5}{10.2}\selectfont
\setlength{\tabcolsep}{2pt}
\renewcommand{\arraystretch}{1.18}
\begin{tabularx}{\linewidth}{@{}Yrrrrrrrr@{}}
\toprule
\textbf{Model} & \textbf{Quality} & \textbf{A--V Sync} & \textbf{Stability} & \textbf{Following} & \textbf{Identity} & \textbf{Q-Overall} & \textbf{Custom.} & \textcolor{tableAccent}{\textbf{Overall}} \\
\midrule
AptAvatar 14B & 55.55 & 98.89 & 78.06 & 13.04 & 81.33 & 65.37 & +5 & 70.37 \\
Avatar-Forever 22B & 34.73 & 34.67 & 45.31 & 6.37 & 74.82 & 39.18 & +5 & 44.18 \\
AvatarForcing 1.3B & 54.32 & 18.44 & 71.60 & 8.44 & 68.50 & 44.26 & +5 & 49.26 \\
LiveTalk 1.3B & 71.27 & 86.44 & 96.06 & 16.59 & 83.89 & 70.85 & +5 & 75.85 \\
Quark Live Avatar 14B & 43.51 & 98.89 & 55.30 & 11.70 & 79.75 & 57.83 & +5 & 62.83 \\
SoulX-LiveAct 18B & 48.39 & 48.67 & 64.94 & 10.74 & 69.83 & 48.51 & +5 & 53.51 \\
Vidu S2 Avatar & 63.68 & 99.60 & 97.25 & 36.90 & 86.12 & 76.71 & +5 & 81.71 \\
\bottomrule
\end{tabularx}
\par\vspace{2pt}{\scriptsize Q-Overall: equal-weight mean of the content metrics. Custom.: sweet-spot bonus, +5 for systems that support a custom avatar and 0 otherwise. Overall $=$ Q-Overall $+$ Custom.}\par\vspace{3pt}
\end{table}

\FloatBarrier
\subsubsection{World Generation}
\label{sec:results_general}

World generation ranks text- and image-conditioned configurations together; a model evaluated with both inputs is reported with its text-conditioned result (\cref{tab:results_general_interaction}). MemFlow achieves the highest Overall score at 33.36, closely followed by Rolling Forcing~\citep{liu2025rollingforcing} at 33.01, Krea Realtime at 31.61, HY-World 1.5 WorldPlay at 31.44, and HappyOyster 1.0 Adventure at 30.84. HappyOyster has the highest Quality (58.34) and Stability (56.67) but low Following (5.56) and Continuity (2.78). Matrix-Game 3.5 achieves the highest Continuity at 34.50, followed by MemFlow at 29.54.

The ordering of models changes across visual quality, continuity, and Following. Krea Realtime records the highest Following score at 25.93, followed by Reward Forcing at 23.33 and MemFlow at 22.59, whereas Rolling Forcing scores 17.78 and Causal Forcing 15.23. Reward Forcing reaches 23.33 in Following but only 16.06 in Stability. Similarly, ShotStream attains 33.80 in Quality but 20.55 in Stability and 9.34 in Continuity (21.02 Overall), underscoring the joint requirements of visual synthesis, temporal continuity, and responsive online control.

% Core world generation results; the inapplicable Reference column is omitted.
\begin{table}[htb]
\centering
\caption{\textbf{World generation.} Text- and image-conditioned configurations are ranked together; a model evaluated with both inputs is reported with its text-conditioned result.}
\label{tab:results_general_interaction}
\definecolor{tableAccent}{HTML}{365F91}
\fontsize{8.5}{10.2}\selectfont
\setlength{\tabcolsep}{3pt}
\renewcommand{\arraystretch}{1.18}
\begin{tabularx}{\linewidth}{@{}Yrrrrr@{}}
\toprule
\textbf{Model} & \textbf{Quality} & \textbf{Stability} & \textbf{Following} & \textbf{Continuity} & \textcolor{tableAccent}{\textbf{Overall}} \\
\midrule
AlayaWorld & 49.44 & 46.17 & 8.89 & 10.08 & 28.65 \\
Causal Forcing & 18.12 & 15.83 & 15.23 & 14.11 & 15.82 \\
Context Forcing & 35.70 & 31.04 & 18.52 & 16.49 & 25.44 \\
HappyOyster 1.0 Adventure & 58.34 & 56.67 & 5.56 & 2.78 & 30.84 \\
Helios & 28.79 & 28.04 & 14.44 & 19.90 & 22.79 \\
HY-World 1.5 WorldPlay & 50.60 & 48.06 & 10.00 & 17.11 & 31.44 \\
Krea Realtime & 35.61 & 42.79 & 25.93 & 22.12 & 31.61 \\
LingBot-World 2.0 & 16.76 & 17.44 & 11.11 & 15.33 & 15.16 \\
LongLive 2.0 & 30.08 & 31.84 & 17.41 & 20.74 & 25.02 \\
Matrix-Game 3.5 & 40.43 & 37.86 & 6.67 & 34.50 & 29.87 \\
MemFlow & 40.93 & 40.38 & 22.59 & 29.54 & 33.36 \\
Reward Forcing & 21.73 & 16.06 & 23.33 & 21.41 & 20.63 \\
Rolling Forcing & 45.19 & 44.47 & 17.78 & 24.61 & 33.01 \\
Self-Forcing & 16.29 & 17.06 & 15.19 & 13.62 & 15.54 \\
ShotStream & 33.80 & 20.55 & 20.37 & 9.34 & 21.02 \\
Yume 1.5 & 25.97 & 27.68 & 19.63 & 26.22 & 24.88 \\
Zing-0.5 & 35.60 & 39.15 & 17.41 & 23.66 & 28.96 \\
\bottomrule
\end{tabularx}
\par\vspace{3pt}
\end{table}

% \FloatBarrier

\Needspace{9\baselineskip}
\subsubsection{Instruction Editing}
\label{sec:results_instruction}

Vidu S2 Editing (45.43 Overall) and JoyAI-Video-Edit~\citep{xiao2026joyaivideoedit} (42.39 Overall) lead instruction editing by a wide margin over the remaining four reported models (\cref{tab:results_instruction_edit}). Vidu S2 Editing achieves the highest scores in Following (72.10), Edit Success (57.20), Quality (22.99), and Stability (13.94), accompanied by 60.93 in Preservation. JoyAI-Video-Edit ranks second in Following (63.89), Edit Success (43.52), and Stability (9.57), while leading all systems in Preservation at 83.33. Lucy 2.5 ranks third at 28.59 Overall, with 36.48 Following, 12.12 Edit Success, and 65.00 Preservation. SANA-Streaming ranks fourth at 16.95 Overall (15.15 Quality, 44.26 Preservation, 4.63 Edit Success), and the other two reported models score at or below 10.29 Overall.

The distribution of metrics across both leading and mid-tier systems highlights the multi-constraint structure of streaming instruction editing. While Following assesses the requested operation and target, Edit Success evaluates complete editing requirements including scope, timing, persistence, and protected-content requirements (Following / Edit Success: 72.10 / 57.20 for Vidu S2 Editing, 63.89 / 43.52 for JoyAI-Video-Edit, and 12.04 / 4.26 for EditaLive (14B)). Meanwhile, Quality (up to 22.99) and Stability (up to 13.94) remain low even for the highest-scoring systems, indicating the difficulty of sustaining clean temporal rendering during live edits.

\begin{table}[htb]
\centering
\caption{\textbf{Instruction editing.}}
\label{tab:results_instruction_edit}
\definecolor{tableAccent}{HTML}{365F91}
\fontsize{8.5}{10.2}\selectfont
\setlength{\tabcolsep}{3pt}
\renewcommand{\arraystretch}{1.18}
\begin{tabularx}{\linewidth}{@{}Yrrrrrr@{}}
\toprule
\textbf{Model} & \textbf{Quality} & \textbf{Stability} & \textbf{Following} & \textbf{Edit Success} & \textbf{Preservation} & \textcolor{tableAccent}{\textbf{Overall}} \\
\midrule
EditaLive (14B) & 7.86 & 4.12 & 12.04 & 4.26 & 23.15 & 10.29 \\
JoyAI-Video-Edit & 11.63 & 9.57 & 63.89 & 43.52 & 83.33 & 42.39 \\
Lucy 2.5 & 19.94 & 9.40 & 36.48 & 12.12 & 65.00 & 28.59 \\
SANA-Streaming & 15.15 & 9.22 & 11.48 & 4.63 & 44.26 & 16.95 \\
Vidu S2 Editing & 22.99 & 13.94 & 72.10 & 57.20 & 60.93 & 45.43 \\
XMax X2.0 & 5.14 & 0.97 & 11.30 & 2.59 & 28.52 & 9.70 \\
\bottomrule
\end{tabularx}
\par\vspace{3pt}
\end{table}

\Needspace{9\baselineskip}
\subsubsection{Reference Editing}
\label{sec:results_reference}

Vidu S2 Editing achieves the highest Overall score in reference editing at 39.02, leading in Reference (31.20), Preservation (64.44), Stability (46.74), Quality (37.72), and Edit Success (15.00) (\cref{tab:results_reference_edit}). Lucy 2.5 ranks second at 18.67 Overall, with 25.55 Quality, 28.55 Stability, 7.59 Reference, 1.30 Edit Success, and 30.38 Preservation. JoyAI-Video-Edit follows at 14.33 Overall and XMax X2.0 at 9.47. On checkpoint-level Reference matching, JoyAI-Video-Edit (26.11) and XMax X2.0 (20.56) follow Vidu S2 Editing (31.20).

Comparing Reference, Preservation, and Edit Success illustrates the distinct requirements of adopting reference appearances, preserving source content, and completing an edit. Although JoyAI-Video-Edit and XMax X2.0 exceed 20 in Reference, their lower Preservation (26.67 and 10.00) and Stability (3.46 and 7.51) accompany Edit Success scores of 8.89 and 1.11, respectively. Conversely, Vidu S2 Editing pairs 64.44 Preservation with the top Edit Success score of 15.00, showing that satisfying the full reference-editing criteria remains challenging across the four reported systems; \cref{sec:analysis} illustrates these requirements in a recorded session.

\begin{table}[htb]
\centering
\caption{\textbf{Reference editing.}}
\label{tab:results_reference_edit}
\definecolor{tableAccent}{HTML}{365F91}
\fontsize{8.5}{10.2}\selectfont
\setlength{\tabcolsep}{3pt}
\renewcommand{\arraystretch}{1.18}
\begin{tabularx}{\linewidth}{@{}Yrrrrrr@{}}
\toprule
\textbf{Model} & \textbf{Quality} & \textbf{Stability} & \textbf{Reference} & \textbf{Edit Success} & \textbf{Preservation} & \textcolor{tableAccent}{\textbf{Overall}} \\
\midrule
JoyAI-Video-Edit & 6.54 & 3.46 & 26.11 & 8.89 & 26.67 & 14.33 \\
Lucy 2.5 & 25.55 & 28.55 & 7.59 & 1.30 & 30.38 & 18.67 \\
Vidu S2 Editing & 37.72 & 46.74 & 31.20 & 15.00 & 64.44 & 39.02 \\
XMax X2.0 & 8.20 & 7.51 & 20.56 & 1.11 & 10.00 & 9.47 \\
\bottomrule
\end{tabularx}
\par\vspace{3pt}
\end{table}

\FloatBarrier
\Needspace{6\baselineskip}
\section{Analysis and Qualitative Results}
\label{sec:analysis}

We examine a recorded reference-editing session to illustrate the divergence between source preservation and reference adoption observed in \cref{sec:results_reference}, and summarize benchmark coverage and evaluation scope.

\FloatBarrier
\Needspace{11\baselineskip}
\subsection{A Recorded Reference-Editing Session}

\begin{figure}[!htb]
  \centering
  \includegraphics[width=\linewidth]{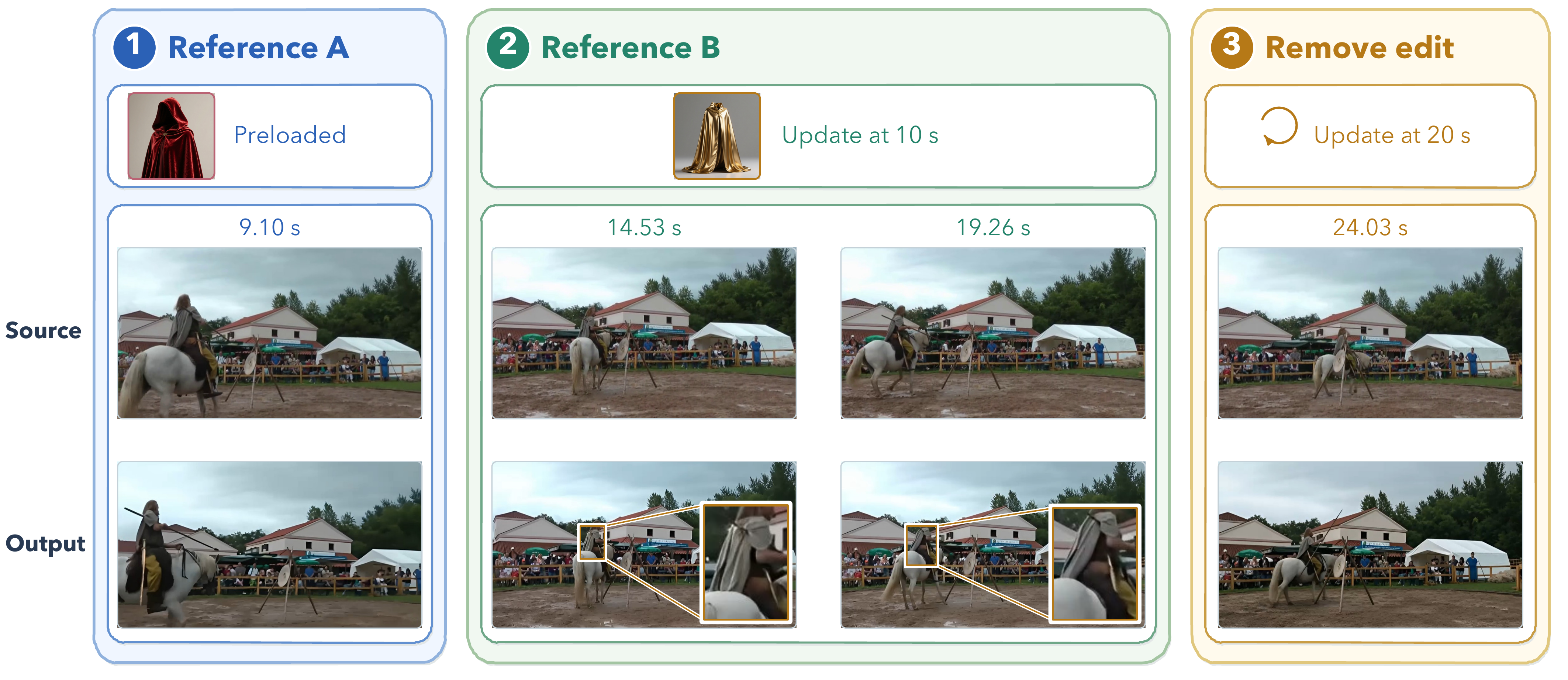}
  \caption{\textbf{Reference editing under changing inputs.} Reference A is preloaded, B arrives at 10\,s, and removal is requested at 20\,s. Columns pair source and output frames at recorded session times. Insets magnify the outlined cape regions, showing grey retained in both B-phase outputs instead of the \mbox{requested gold satin.}}
  \label{fig:editing_case}
\end{figure}

\Cref{fig:editing_case} compares source preservation and reference adoption in a recorded session.\footnote{Source footage: \href{https://commons.wikimedia.org/wiki/File:Villa-borg-2011-reitkunst.ogv}{Rhmaster}, \href{https://creativecommons.org/licenses/by-sa/3.0/}{CC BY-SA 3.0}.} The prompt schedule requests a red velvet cape (Reference A) initially, switches to a gold satin cape (Reference B) at 10\,s, and requests removal at 20\,s. Both sampled B-phase outputs pair the generated video with the latest contemporaneous source frame: the models preserve the underlying scene, but retain the original grey cape rather than adopting the requested gold satin appearance.

% \FloatBarrier
% \Needspace{9\baselineskip}
\subsection{Benchmark Coverage}

The benchmark combines synthetic portraits and speech, controlled acoustic variations, and licensed source videos. Its editing tasks test reference switching, multi-target retention, and restoration against an advancing source video, while future extensions can cover broader identities and environments, global resets, and selective restoration while concurrent edits remain active.
\FloatBarrier
\Needspace{10\baselineskip}
\section{Related Work}
\label{sec:related_work}

\subsection{Streaming Video Models}
\label{sec:related_models}

\subsubsection{Streaming Video Generation}
\paragraph{Efficient causal generation.}
Incremental video synthesis requires efficient sampling and temporal models that can use previously generated content. StreamDiffusion develops efficient pipelines for interactive image synthesis~\citep{kodaira2023streamdiffusion}, while Diffusion Forcing provides a flexible formulation for sequence prediction~\citep{chen2024diffusionforcing}. CausVid brings few-step distillation to causal video generation~\citep{yin2024causvid}; Self Forcing and Causal Forcing further address the mismatch between training and autoregressive inference~\citep{huang2025selfforcing,zhu2026causalforcing}. Causal Forcing++ further studies frame-wise generation with one or two sampling steps~\citep{zhao2026causalforcingpp}. rCM scales consistency distillation to large image and video models, and Causal-rCM extends it to autoregressive streaming diffusion~\citep{zheng2026rcm,zheng2026causalrcm}. Speculative decoding for video generation uses a smaller drafter and a quality router to reduce target-model computation~\citep{hu2026sdvg}. StreamDiT and StreamDiffusionV2 develop continuous generation and serving pipelines~\citep{kodaira2025streamdit,feng2025streamdiffusionv2}, making incremental synthesis practical for interactive applications.

\paragraph{Efficient inference and serving.}
Efficient attention methods span hardware-efficient kernels, sparsity, compact representations, and linear formulations~\citep{zhang2025efficientattention}. SageAttention and its successors accelerate low-bit inference~\citep{zhang2025sageattention,zhang2025sageattention2,zhang2025sageattention2pp,zhang2025sageattention3}, while SageBwd studies low-bit training~\citep{zhang2026sagebwd}; dynamic block-level INT8 fallback handles training-time activation outliers~\citep{zhang2025int8fallback}. SpargeAttention and SpargeAttention2 study sparse attention, while SLA and SLA2 combine sparse and linear computation~\citep{zhang2025spargeattention,zhang2026spargeattention2,zhang2026sla,zhang2026sla2}. Sparse VideoGen and Sparse VideoGen2 exploit spatial-temporal sparsity and semantic grouping, and SVG-EAR adds error-aware routing and linear compensation~\citep{xi2025sparsevideogen,yang2025sparsevideogen2,zhou2026svgear}. Mixed-precision quantization and diffusion distillation further reduce video-generation cost~\citep{su2026sixbitdiffusion,zhang2025turbodiffusion}. Related image-generation work studies pixel- and timestep-level model stitching for diffusion acceleration~\citep{sun2026hybridstitch}.

Flash-KMeans targets online clustering for token grouping and KV-cache quantization~\citep{yang2026flashkmeans}. KernelBenchX evaluates correctness and hardware efficiency of LLM-generated GPU kernels~\citep{wang2026kernelbenchx}. TurboServe addresses stateful-session scheduling and GPU provisioning~\citep{jiang2026turboserve}; Vidu S1 and S2 connect efficient synthesis to live character interaction and streaming editing~\citep{zhang2026vidu,zhang2026vidus2}. These implementation advances motivate measuring client-observed response and sustained delivery alongside content quality.

\paragraph{Memory and changing controls.}
Sustained generation adds the problem of retaining useful history as the stream grows. FramePack bounds the context cost by compressing historical frames~\citep{zhang2025framepack}; Rolling Forcing, Reward Forcing, and Helios address accumulated drift and motion degradation over long rollouts~\citep{liu2025rollingforcing,lu2025rewardforcing,yuan2026helios}. Context Forcing trains with longer historical context, while MemFlow retrieves relevant past frames~\citep{chen2026contextforcing,ji2025memflow}. Changing requests also require deciding which earlier states remain relevant. LongLive supports prompt transitions during generation~\citep{yang2025longlive}, and Yume-1.5 adds text-controlled world events~\citep{mao2025yume15}. IAMFlow tracks persistent entities across narrative updates~\citep{liu2026iamflow}, and SEGUE models the intermediate state changes needed to execute a prompt switch~\citep{rao2026segue}. Quant VideoGen compresses autoregressive KV caches, UltraViCo addresses video-length extrapolation, and ViewRoPE introduces camera-ray geometry for spatial persistence~\citep{xi2026quantvideogen,zhao2026ultravico,xiang2026viewrope}. These developments connect long-term memory with selective adaptation to new controls.

\paragraph{Avatars and spoken interaction.}
Avatar generation applies these capabilities to speech, identity, and bodily motion. VASA-1 animates portraits from supplied audio, MuseTalk focuses on identity-preserving lip synchronization~\citep{zhang2025musetalk}, and OmniAvatar extends audio-driven synthesis to text-controlled body animation~\citep{xu2024vasa1,gan2025omniavatar}. Live Avatar, SoulX-FlashTalk, and AvatarForcing pursue efficient, sustained avatar rendering~\citep{huang2025liveavatar,shen2025soulxflashtalk,cui2026avatarforcing}. SoulX-LiveAct extends sustained human animation with compressed historical memory~\citep{zhen2026soulxliveact}. Joint audiovisual models also generate both modalities during streaming, as in OmniForcing and Omni-LiveAvatar~\citep{su2026omniforcing,zhu2026omniliveavatar}. TurboT2VA studies efficient joint audio-video synthesis through consistency distillation~\citep{yang2026turbot2va}. Conversational interaction additionally requires deciding what to say and when to respond, drawing on real-time speech dialogue in Moshi and multimodal interactive synthesis in LiveTalk~\citep{defossez2024moshi,chern2025livetalk}. This progression motivates separate evaluation of prescribed avatar performance and dialogue behavior.

\subsubsection{Streaming Video Editing}
\paragraph{Preserving an advancing source.}
Video editing introduces an external source whose content and motion constrain the generated output. InstructPix2Pix establishes instruction-conditioned image editing, while TokenFlow addresses consistency across edited video frames~\citep{brooks2022instructpix2pix,geyer2023tokenflow}. StreamV2V and Live2Diff support incremental processing through historical features and causal attention~\citep{liang2024streamv2v,xing2024live2diff}. Recent work combines efficient inference with stronger editing and preservation: LiveEdit and SANA-Streaming develop few-step streaming editors~\citep{wang2026liveedit,zhao2026sanastreaming}, StreamEdit adapts streaming generators without additional training~\citep{jiao2026streamedit}, and EgoEdit targets egocentric scenes with hand--object interactions~\citep{li2025egoedit}. JoyAI-Video-Edit further addresses source drift during sustained autoregressive editing~\citep{xiao2026joyaivideoedit}.

\paragraph{Instructions and visual references.}
Editing controls specify both the desired appearance and the content to preserve. VACE unifies images, source videos, and masks as conditioning inputs, with an autoregressive adaptation enabling streaming use~\citep{jiang2025vace,fosdick2026streamingvace}. EditaLive transfers appearance from edited references while retaining character motion~\citep{li2026editalive}, and Vidu S2-Editing combines source and reference conditioning for streaming edits~\citep{zhang2026vidus2}. These interfaces support different editing requests. In \ours, successive instructions or references specify when an edit should take effect, which established changes should persist, and when edited content should return to the current source.

\subsection{Evaluating Streaming Video Systems}
\label{sec:related_evaluation}
\paragraph{Quality, semantics, and automated judging.}
Video evaluation has progressed from distributional measures such as Fr\'echet Video Distance~\citep{unterthiner2018fvd} to decomposed assessments of visual quality, motion, and semantic fidelity in VBench, VBench++, and EvalCrafter~\citep{huang2024vbench,huang2024vbenchpp,liu2024evalcrafter}. T2V-CompBench and GenAI-Bench examine compositional alignment~\citep{sun2024t2vcompbench,li2024genaibench}, while VBench-2.0 broadens evaluation to intrinsic faithfulness, including physics and commonsense~\citep{zheng2025vbench2}. Video-Bench studies human-aligned judging of generated videos~\citep{han2025videobench}; MLLM-as-a-Judge examines multimodal judges across scoring, comparison, and ranking~\citep{chen2024mllmjudge}. These works provide content-level evaluation foundations and motivate checking the reliability of automated judgments.

\paragraph{Temporal change and long-term memory.}
Evaluating evolving content requires examining both intended changes and retained state. TC-Bench assesses the completion of transitions between specified scene states~\citep{feng2024tcbench}. NarraStream-Bench evaluates entity grounding, cross-segment consistency, and instruction following under multiple prompts~\citep{liu2026iamflow}; MBench examines entity, environment, and causal consistency as aspects of world-model memory~\citep{zhang2026mbench}. OpenTrans-360 further evaluates how earlier states end and new states emerge during prompt switches~\citep{rao2026segue}. ViewBench diagnoses loop-closure fidelity and geometric drift~\citep{xiang2026viewrope}. These benchmarks establish temporal behavior and memory as evaluation targets beyond the quality of individual frames.

\paragraph{Multi-turn and streaming interaction.}
Interactive benchmarks connect generated changes to successive controls. A first-principles account of general world models places such interaction within a broader cycle of understanding, imagination, and action~\citep{zhuGeneralWorldModels}. WorldScore assesses world controllability, quality, and dynamics~\citep{duan2025worldscore}, while WorldMark standardizes scenes and action interfaces across interactive world models~\citep{xu2026worldmark}. WBench evaluates multi-turn navigation, subject actions, event editing, and perspective changes~\citep{ying2026wbench}; its event edits modify the generated world. EditWorld's WBench-Editing examines successive world edits and preservation of unrelated content through a main text-only comparison and a separate reference-conditioned subset~\citep{liao2026editworld}. StreamAV-Bench addresses streaming audio-video generation through progressive and interactive tracks, assessing prompt updates, response realization, and state retention and reuse~\citep{liu2026streamavbench}. Its diagnostics also account for unsuccessful updates. These protocols already capture important aspects of interactive behavior, providing direct points of comparison for \ours.

\paragraph{Avatar performance and dialogue.}
Avatar evaluation adds identity and audiovisual requirements. SyncNet provides a measure of lip synchronization~\citep{chung2016syncnet}, while THEval and Live Avatar's GenBench evaluate rendered speakers, including visual quality and synchronization across different durations~\citep{quignon2025theval,huang2025liveavatar}. Full-Duplex-Bench evaluates conversational behaviors such as pausing, backchanneling, turn-taking, and interruption~\citep{lin2025fullduplexbench}. VideoFDB evaluates full-duplex audiovisual conversation through separate perception and generation assessments of nonverbal cues~\citep{mazumdar2026videofdb}. This emphasis on conversational appropriateness complements our dialogue execution and prescribed-performance tasks, which do not test perception of a live user-video stream. These complementary assessments distinguish the quality of a rendered performance from the behavior of a system participating in a conversation.

\paragraph{Real-time video editing.}
Editing benchmarks assess requested changes together with source preservation. TDVE-DB provides human ratings of quality, alignment, and structural consistency for TDVE-Assessor~\citep{wang2025tdveassessor}. EditVerseBench covers diverse editing operations, including reference insertion and first-frame propagation~\citep{ju2025editverse}. LiveEdit benchmarks streaming editing~\citep{wang2026liveedit}, while EgoEditBench emphasizes egocentric edits and preservation under hand--object interactions~\citep{li2025egoedit}. CharEdit-Bench evaluates character transformations over short and long videos~\citep{li2026editalive}, and LongV2VBench tests global and local editing operations over long videos~\citep{xiao2026joyaivideoedit}. These studies make source retention and long-horizon editing quality important complements to edit alignment.

\ours brings these evaluation requirements together across dialogue, prescribed avatar performance, world generation, instruction editing, and reference editing. Each task specifies when inputs become available, what should change, and which state or source content should remain. The resulting assessment links task execution and state consistency with visual fidelity and audiovisual coordination. \Cref{sec:benchmark_comparison} compares five-task coverage and evaluation duration across existing benchmarks.

% \Needspace{16\baselineskip}
\section{Conclusion}
\label{sec:conclusion}

\ours brings five interactive streaming video tasks into a unified benchmark of 300 scripted test cases, linking evolving requests to explicit execution and state-preservation requirements. Our evaluation of 38 models reveals complementary strengths across tasks, while online action following and reference-based editing remain challenging. In reference editing, the highest reported Edit Success score is 15.00 on a 0--100 scale. Improving request execution alongside visual fidelity and state continuity is therefore an important direction for future model design. A natural next step is to collect human pairwise judgments and examine their agreement with the task-specific evaluation rubric. Together, these directions support progress toward omni-modal real-time interactive applications that integrate continuous perception, generation, and responsive control.

\FloatBarrier
% Broader background in retrieval, language-model serving, diffusion, and quantization.
% Explicit keys include the remaining non-database works from Scholar profile J3NTnmEAAAAJ.
\nocite{zhang2025sagerag,jiang2025cascadia,hu2026residualcontext,zhao2025ultraimage,hu2025postquantization,jiang2026hexgen3}
\bibliographystyle{plainnat}
\bibliography{main}

@article{ju2025editverse,
  title   = {{EditVerse: Unifying Image and Video Editing and Generation with In-Context Learning}},
  author  = {Ju, Xuan and Wang, Tianyu and Zhou, Yuqian and Zhang, He and Liu, Qing and Zhao, Nanxuan and Zhang, Zhifei and Li, Yijun and Cai, Yuanhao and Liu, Shaoteng and Pakhomov, Daniil and Lin, Zhe and Kim, Soo Ye and Xu, Qiang},
  journal = {arXiv preprint arXiv:2509.20360},
  year    = {2025},
  url     = {https://arxiv.org/abs/2509.20360}
}

@article{wang2026liveedit,
  title   = {{LiveEdit: Towards Real-Time Diffusion-Based Streaming Video Editing}},
  author  = {Wang, Xinyu and Zhao, Chongbo and Zhan, Fangneng and Ma, Yue},
  journal = {arXiv preprint arXiv:2606.26740},
  year    = {2026},
  url     = {https://arxiv.org/abs/2606.26740}
}

@article{zhang2026vidu,
  title = {{Vidu S1: A Real-Time Interactive Video Generation Model}},
  author = {Zhang, Jintao and Jiang, Kai and Chen, Jintao and Wang, Xu and Luo, Yang and Wang, Yuji and Chen, Dechuang and Li, Jungang and Ye, Chengyang and Chen, Marco and Zhu, Hongzhou and Zhao, Min and Jiang, Yuxuan and Huang, Zhengkun and Xiang, Chendong and Zheng, Kaiwen and Wang, Haoxu and Wang, Xiaohang and Jia, Qi and Chen, Xin and Chen, Yimin and Jiang, Youhe and Fu, Fangcheng and Deng, Zhijie and Bao, Fan and Chen, Jianfei and Zhu, Jun},
  journal = {arXiv preprint arXiv:2607.03118},
  year = {2026},
  url = {https://arxiv.org/abs/2607.03118}
}

@article{zhao2026sanastreaming,
  title   = {{SANA-Streaming: Real-time Streaming Video Editing with Hybrid Diffusion Transformer}},
  author  = {Zhao, Yuyang and Pan, Yicheng and He, Qiyuan and Yu, Jincheng and Chen, Junsong and Ye, Tian and Liu, Haozhe and Xie, Enze and Han, Song},
  journal = {arXiv preprint arXiv:2605.30409},
  year    = {2026},
  url     = {https://arxiv.org/abs/2605.30409}
}

@article{jiao2026streamedit,
  title   = {{StreamEdit: Training-Free Video Editing via Few-Step Streaming Video Generation}},
  author  = {Jiao, Guanlong and Zhang, Chenyangguang and Xian, Jia Jun Cheng and Zhang, Zewei and Liao, Renjie},
  journal = {arXiv preprint arXiv:2605.21466},
  year    = {2026},
  url     = {https://arxiv.org/abs/2605.21466}
}

@article{yang2025longlive,
  title   = {{LongLive: Real-time Interactive Long Video Generation}},
  author  = {Yang, Shuai and Huang, Wei and Chu, Ruihang and Xiao, Yicheng and Zhao, Yuyang and Wang, Xianbang and Li, Muyang and Xie, Enze and Chen, Yingcong and Lu, Yao and Han, Song and Chen, Yukang},
  journal = {arXiv preprint arXiv:2509.22622},
  year    = {2025},
  url     = {https://arxiv.org/abs/2509.22622}
}

@article{kodaira2025streamdit,
  title   = {{StreamDiT: Real-Time Streaming Text-to-Video Generation}},
  author  = {Kodaira, Akio and Hou, Tingbo and Hou, Ji and Georgopoulos, Markos and Juefei-Xu, Felix and Tomizuka, Masayoshi and Zhao, Yue},
  journal = {arXiv preprint arXiv:2507.03745},
  year    = {2025},
  url     = {https://arxiv.org/abs/2507.03745}
}

@article{zhu2026causalforcing,
  title   = {{Causal Forcing: Autoregressive Diffusion Distillation Done Right for High-Quality Real-Time Interactive Video Generation}},
  author  = {Zhu, Hongzhou and Zhao, Min and He, Guande and Su, Hang and Li, Chongxuan and Zhu, Jun},
  journal = {arXiv preprint arXiv:2602.02214},
  year    = {2026},
  url     = {https://arxiv.org/abs/2602.02214}
}

@article{huang2024vbenchpp,
  title   = {{VBench++: Comprehensive and Versatile Benchmark Suite for Video Generative Models}},
  author  = {Huang, Ziqi and Zhang, Fan and Xu, Xiaojie and He, Yinan and Yu, Jiashuo and Dong, Ziyue and Ma, Qianli and Chanpaisit, Nattapol and Si, Chenyang and Jiang, Yuming and Wang, Yaohui and Chen, Xinyuan and Chen, Ying-Cong and Wang, Limin and Lin, Dahua and Qiao, Yu and Liu, Ziwei},
  journal = {arXiv preprint arXiv:2411.13503},
  year    = {2024},
  url     = {https://arxiv.org/abs/2411.13503}
}

@article{han2025videobench,
  title   = {{Video-Bench: Human-Aligned Video Generation Benchmark}},
  author  = {Han, Hui and Li, Siyuan and Chen, Jiaqi and Yuan, Yiwen and Wu, Yuling and Leong, Chak Tou and Du, Hanwen and Fu, Junchen and Li, Youhua and Zhang, Jie and Zhang, Chi and Li, Li-jia and Ni, Yongxin},
  journal = {arXiv preprint arXiv:2504.04907},
  year    = {2025},
  url     = {https://arxiv.org/abs/2504.04907}
}

@article{kodaira2023streamdiffusion,
  title   = {{StreamDiffusion: A Pipeline-level Solution for Real-time Interactive Generation}},
  author  = {Kodaira, Akio and Xu, Chenfeng and Hazama, Toshiki and Yoshimoto, Takanori and Ohno, Kohei and Mitsuhori, Shogo and Sugano, Soichi and Cho, Hanying and Liu, Zhijian and Tomizuka, Masayoshi and Keutzer, Kurt},
  journal = {arXiv preprint arXiv:2312.12491},
  year    = {2023},
  url     = {https://arxiv.org/abs/2312.12491}
}

@article{yin2024causvid,
  title   = {{From Slow Bidirectional to Fast Autoregressive Video Diffusion Models}},
  author  = {Yin, Tianwei and Zhang, Qiang and Zhang, Richard and Freeman, William T. and Durand, Fredo and Shechtman, Eli and Huang, Xun},
  journal = {arXiv preprint arXiv:2412.07772},
  year    = {2024},
  url     = {https://arxiv.org/abs/2412.07772}
}

@article{huang2025selfforcing,
  title   = {{Self Forcing: Bridging the Train-Test Gap in Autoregressive Video Diffusion}},
  author  = {Huang, Xun and Li, Zhengqi and He, Guande and Zhou, Mingyuan and Shechtman, Eli},
  journal = {arXiv preprint arXiv:2506.08009},
  year    = {2025},
  url     = {https://arxiv.org/abs/2506.08009}
}

@article{liang2024streamv2v,
  title   = {{Looking Backward: Streaming Video-to-Video Translation with Feature Banks}},
  author  = {Liang, Feng and Kodaira, Akio and Xu, Chenfeng and Tomizuka, Masayoshi and Keutzer, Kurt and Marculescu, Diana},
  journal = {arXiv preprint arXiv:2405.15757},
  year    = {2024},
  url     = {https://arxiv.org/abs/2405.15757}
}

@article{xing2024live2diff,
  title   = {{Live2Diff: Live Stream Translation via Uni-directional Attention in Video Diffusion Models}},
  author  = {Xing, Zhening and Fox, Gereon and Zeng, Yanhong and Pan, Xingang and Elgharib, Mohamed and Theobalt, Christian and Chen, Kai},
  journal = {arXiv preprint arXiv:2407.08701},
  year    = {2024},
  url     = {https://arxiv.org/abs/2407.08701}
}

@article{xu2024vasa1,
  title   = {{VASA-1: Lifelike Audio-Driven Talking Faces Generated in Real Time}},
  author  = {Xu, Sicheng and Chen, Guojun and Guo, Yu-Xiao and Yang, Jiaolong and Li, Chong and Zang, Zhenyu and Zhang, Yizhong and Tong, Xin and Guo, Baining},
  journal = {arXiv preprint arXiv:2404.10667},
  year    = {2024},
  url     = {https://arxiv.org/abs/2404.10667}
}

@article{brooks2022instructpix2pix,
  title   = {{InstructPix2Pix: Learning to Follow Image Editing Instructions}},
  author  = {Brooks, Tim and Holynski, Aleksander and Efros, Alexei A.},
  journal = {arXiv preprint arXiv:2211.09800},
  year    = {2022},
  url     = {https://arxiv.org/abs/2211.09800}
}

@article{geyer2023tokenflow,
  title   = {{TokenFlow: Consistent Diffusion Features for Consistent Video Editing}},
  author  = {Geyer, Michal and Bar-Tal, Omer and Bagon, Shai and Dekel, Tali},
  journal = {arXiv preprint arXiv:2307.10373},
  year    = {2023},
  url     = {https://arxiv.org/abs/2307.10373}
}

@article{defossez2024moshi,
  title   = {{Moshi: a speech-text foundation model for real-time dialogue}},
  author  = {D{\'e}fossez, Alexandre and Mazar{\'e}, Laurent and Orsini, Manu and Royer, Am{\'e}lie and P{\'e}rez, Patrick and J{\'e}gou, Herv{\'e} and Grave, Edouard and Zeghidour, Neil},
  journal = {arXiv preprint arXiv:2410.00037},
  year    = {2024},
  url     = {https://arxiv.org/abs/2410.00037}
}

@article{lin2025fullduplexbench,
  title   = {{Full-Duplex-Bench: A Benchmark to Evaluate Full-duplex Spoken Dialogue Models on Turn-taking Capabilities}},
  author  = {Lin, Guan-Ting and Lian, Jiachen and Li, Tingle and Wang, Qirui and Anumanchipalli, Gopala and Liu, Alexander H. and Lee, Hung-yi},
  journal = {arXiv preprint arXiv:2503.04721},
  year    = {2025},
  url     = {https://arxiv.org/abs/2503.04721}
}

@article{unterthiner2018fvd,
  title   = {{Towards Accurate Generative Models of Video: A New Metric \& Challenges}},
  author  = {Unterthiner, Thomas and van Steenkiste, Sjoerd and Kurach, Karol and Marinier, Raphael and Michalski, Marcin and Gelly, Sylvain},
  journal = {arXiv preprint arXiv:1812.01717},
  year    = {2018},
  url     = {https://arxiv.org/abs/1812.01717}
}

@inproceedings{huang2024vbench,
  title = {{VBench}: Comprehensive Benchmark Suite for Video Generative Models},
  author = {Huang, Ziqi and He, Yinan and Yu, Jiashuo and Zhang, Fan and Si, Chenyang and Jiang, Yuming and Zhang, Yuanhan and Wu, Tianxing and Jin, Qingyang and Chanpaisit, Nattapol and Wang, Yaohui and Chen, Xinyuan and Wang, Limin and Lin, Dahua and Qiao, Yu and Liu, Ziwei},
  booktitle = {Proceedings of the IEEE/CVF Conference on Computer Vision and Pattern Recognition},
  year = {2024},
  url = {https://openaccess.thecvf.com/content/CVPR2024/html/Huang_VBench_Comprehensive_Benchmark_Suite_for_Video_Generative_Models_CVPR_2024_paper.html},
  pages = {21807--21818}
}

@article{liu2024evalcrafter,
  title   = {{EvalCrafter: Benchmarking and Evaluating Large Video Generation Models}},
  author  = {Liu, Yaofang and Cun, Xiaodong and Liu, Xuebo and Wang, Xintao and Zhang, Yong and Chen, Haoxin and Liu, Yang and Zeng, Tieyong and Chan, Raymond and Shan, Ying},
  journal = {arXiv preprint arXiv:2310.11440},
  year    = {2023},
  url     = {https://arxiv.org/abs/2310.11440}
}

@article{sun2024t2vcompbench,
  title   = {{T2V-CompBench: A Comprehensive Benchmark for Compositional Text-to-video Generation}},
  author  = {Sun, Kaiyue and Huang, Kaiyi and Liu, Xian and Wu, Yue and Xu, Zihan and Li, Zhenguo and Liu, Xihui},
  journal = {arXiv preprint arXiv:2407.14505},
  year    = {2024},
  url     = {https://arxiv.org/abs/2407.14505}
}

@article{liu2026streamavbench,
  title   = {{StreamAV-Bench: A Comprehensive Benchmark for Streaming Audio-Video Generation}},
  author  = {Liu, Kaiqi and Zeng, Haoxuan and Liu, Jingqi and Fang, Jiacong and Cai, Ziqi and Mao, Yunyao and Liu, Henglin and Sheng, Yu and Weng, Shuchen and Shi, Boxin},
  journal = {arXiv preprint arXiv:2608.26336},
  year    = {2026},
  url     = {https://arxiv.org/abs/2608.26336}
}

@article{chen2024mllmjudge,
  title   = {{MLLM-as-a-Judge: Assessing Multimodal LLM-as-a-Judge with Vision-Language Benchmark}},
  author  = {Chen, Dongping and Chen, Ruoxi and Zhang, Shilin and Liu, Yinuo and Wang, Yaochen and Zhou, Huichi and Zhang, Qihui and Wan, Yao and Zhou, Pan and Sun, Lichao},
  journal = {arXiv preprint arXiv:2402.04788},
  year    = {2024},
  url     = {https://arxiv.org/abs/2402.04788}
}

@inproceedings{chung2016syncnet,
  title     = {Out of Time: Automated Lip Sync in the Wild},
  author    = {Chung, Joon Son and Zisserman, Andrew},
  booktitle = {Workshop on Multi-view Lip-reading, Asian Conference on Computer Vision},
  year      = {2016},
  url       = {https://www.robots.ox.ac.uk/~vgg/publications/2016/Chung16a/}
}

@article{chen2024diffusionforcing,
  title   = {{Diffusion Forcing: Next-token Prediction Meets Full-Sequence Diffusion}},
  author  = {Chen, Boyuan and Monso, Diego Marti and Du, Yilun and Simchowitz, Max and Tedrake, Russ and Sitzmann, Vincent},
  journal = {arXiv preprint arXiv:2407.01392},
  year    = {2024},
  url     = {https://arxiv.org/abs/2407.01392}
}

@article{zhang2025framepack,
  title   = {{Frame Context Packing and Drift Prevention in Next-Frame-Prediction Video Diffusion Models}},
  author  = {Zhang, Lvmin and Cai, Shengqu and Li, Muyang and Wetzstein, Gordon and Agrawala, Maneesh},
  journal = {arXiv preprint arXiv:2504.12626},
  year    = {2025},
  url     = {https://arxiv.org/abs/2504.12626}
}

@article{yuan2026helios,
  title   = {{Helios: Real Real-Time Long Video Generation Model}},
  author  = {Yuan, Shenghai and Yin, Yuanyang and Li, Zongjian and Huang, Xinwei and Yang, Xiao and Yuan, Li},
  journal = {arXiv preprint arXiv:2603.04379},
  year    = {2026},
  url     = {https://arxiv.org/abs/2603.04379}
}

@article{gan2025omniavatar,
  title   = {{OmniAvatar: Efficient Audio-Driven Avatar Video Generation with Adaptive Body Animation}},
  author  = {Gan, Qijun and Yang, Ruizi and Zhu, Jianke and Xue, Shaofei and Hoi, Steven},
  journal = {arXiv preprint arXiv:2506.18866},
  year    = {2025},
  url     = {https://arxiv.org/abs/2506.18866}
}

@article{huang2025liveavatar,
  title   = {{Live Avatar: Streaming Real-time Audio-Driven Avatar Generation with Infinite Length}},
  author  = {Huang, Yubo and Guo, Hailong and Wu, Fangtai and Wang, Weiqiang and Zhang, Shifeng and Huang, Shijie and Gan, Qijun and Liu, Lin and Zhao, Sirui and Chen, Enhong and Liu, Jiaming and Hoi, Steven},
  journal = {arXiv preprint arXiv:2512.04677},
  year    = {2025},
  url     = {https://arxiv.org/abs/2512.04677}
}

@article{shen2025soulxflashtalk,
  title = {{SoulX-FlashTalk: Real-Time Infinite Streaming of Audio-Driven Avatars via Self-Correcting Bidirectional Distillation}},
  author = {Shen, Le and Qiao, Qian and Yu, Tan and Zhou, Ke and Yu, Tianhang and Zhan, Yu and Wang, Zhenjie and Tao, Ming and Yin, Shunshun and Liu, Siyuan},
  journal = {arXiv preprint arXiv:2512.23379},
  year = {2025},
  url = {https://arxiv.org/abs/2512.23379}
}

@article{zhu2026omniliveavatar,
  title   = {{Omni-LiveAvatar: Minute-Level Real-Time Streaming Joint Audio-Video Avatar Generation}},
  author  = {Zhu, Lunjie and Ge, Xingtong and Lin, Fangyu and Zhang, Yi and Liu, Zhening and Li, Mengfei and Zhang, Yumeng and Song, Guanglu and Liu, Yu and Zhang, Jun},
  journal = {arXiv preprint arXiv:2608.13602},
  year    = {2026},
  url     = {https://arxiv.org/abs/2608.13602}
}

@article{chern2025livetalk,
  title   = {{LiveTalk: Real-Time Multimodal Interactive Video Diffusion via Improved On-Policy Distillation}},
  author  = {Chern, Ethan and Hu, Zhulin and Tang, Bohao and Su, Jiadi and Chern, Steffi and Deng, Zhijie and Liu, Pengfei},
  journal = {arXiv preprint arXiv:2512.23576},
  year    = {2025},
  url     = {https://arxiv.org/abs/2512.23576}
}

@article{jiang2025vace,
  title   = {{VACE: All-in-One Video Creation and Editing}},
  author  = {Jiang, Zeyinzi and Han, Zhen and Mao, Chaojie and Zhang, Jingfeng and Pan, Yulin and Liu, Yu},
  journal = {arXiv preprint arXiv:2503.07598},
  year    = {2025},
  url     = {https://arxiv.org/abs/2503.07598}
}

@article{fosdick2026streamingvace,
  title   = {{Adapting VACE for Real-Time Autoregressive Video Diffusion}},
  author  = {Fosdick, Ryan},
  journal = {arXiv preprint arXiv:2602.14381},
  year    = {2026},
  url     = {https://arxiv.org/abs/2602.14381}
}

@article{li2026editalive,
  title   = {{EditaLive! Unified Character Video Editing for Live Streaming}},
  author  = {Li, Zhiyuan and Pun, Chi-Man and Jiang, Peng-Tao and Li, Bo and Cun, Xiaodong},
  journal = {arXiv preprint arXiv:2608.27123},
  year    = {2026},
  url     = {https://arxiv.org/abs/2608.27123}
}

@article{li2024genaibench,
  title   = {{GenAI-Bench: Evaluating and Improving Compositional Text-to-Visual Generation}},
  author  = {Li, Baiqi and Lin, Zhiqiu and Pathak, Deepak and Li, Jiayao and Fei, Yixin and Wu, Kewen and Ling, Tiffany and Xia, Xide and Zhang, Pengchuan and Neubig, Graham and Ramanan, Deva},
  journal = {arXiv preprint arXiv:2406.13743},
  year    = {2024},
  url     = {https://arxiv.org/abs/2406.13743}
}

@article{zheng2025vbench2,
  title   = {{VBench-2.0: Advancing Video Generation Benchmark Suite for Intrinsic Faithfulness}},
  author  = {Zheng, Dian and Huang, Ziqi and Liu, Hongbo and Zou, Kai and He, Yinan and Zhang, Fan and Gu, Lulu and Zhang, Yuanhan and He, Jingwen and Zheng, Wei-Shi and Qiao, Yu and Liu, Ziwei},
  journal = {arXiv preprint arXiv:2503.21755},
  year    = {2025},
  url     = {https://arxiv.org/abs/2503.21755}
}

@article{duan2025worldscore,
  title   = {{WorldScore: A Unified Evaluation Benchmark for World Generation}},
  author  = {Duan, Haoyi and Yu, Hong-Xing and Chen, Sirui and Fei-Fei, Li and Wu, Jiajun},
  journal = {arXiv preprint arXiv:2504.00983},
  year    = {2025},
  url     = {https://arxiv.org/abs/2504.00983}
}

@article{wang2025tdveassessor,
  title   = {{TDVE-Assessor: Benchmarking and Evaluating the Quality of Text-Driven Video Editing with LMMs}},
  author  = {Wang, Juntong and Wang, Jiarui and Duan, Huiyu and Zhai, Guangtao and Min, Xiongkuo},
  journal = {arXiv preprint arXiv:2505.19535},
  year    = {2025},
  url     = {https://arxiv.org/abs/2505.19535}
}

@article{quignon2025theval,
  title   = {{THEval. Evaluation Framework for Talking Head Video Generation}},
  author  = {Quignon, Nabyl and Chopin, Baptiste and Wang, Yaohui and Dantcheva, Antitza},
  journal = {arXiv preprint arXiv:2511.04520},
  year    = {2025},
  url     = {https://arxiv.org/abs/2511.04520}
}

@article{feng2025streamdiffusionv2,
  title   = {{StreamDiffusionV2: A Streaming System for Dynamic and Interactive Video Generation}},
  author  = {Feng, Tianrui and Li, Zhi and Yang, Shuo and Xi, Haocheng and Li, Muyang and Li, Xiuyu and Zhang, Lvmin and Yang, Keting and Peng, Kelly and Han, Song and Agrawala, Maneesh and Keutzer, Kurt and Kodaira, Akio and Xu, Chenfeng},
  journal = {arXiv preprint arXiv:2511.07399},
  year    = {2025},
  url     = {https://arxiv.org/abs/2511.07399}
}

@article{xiao2026joyaivideoedit,
  title   = {{JoyAI-Video-Edit: Real-Time Open-Ended Video Editing with Autoregressive Diffusion}},
  author  = {Xiao, Yicheng and Dai, Wenxun and Qin, Xinran and Song, Lin and Zhang, Maoquan and Xu, Hang and Chen, Yukang and Li, Yitong and Zhang, Guohui and Zhang, Yuan and Zhang, Xuying and Zhang, Tommy and Yuan, Jianlong and Li, Peihao and Lu, Shuai and Fu, Siming and Zhao, Chuyang and Han, Xin and Huang, Jie and Li, Wenbo and Ma, Guoqing and Huang, Wei and Qi, Xiaojuan and Huang, Haoyang and Duan, Nan},
  journal = {arXiv preprint arXiv:2608.03974},
  year    = {2026},
  url     = {https://arxiv.org/abs/2608.03974}
}

@article{zhang2026vidus2,
  title = {{Vidu S2: Real-Time Interactive, Editable, and Spatial Video Generation}},
  author = {Zhang, Jintao and Jiang, Kai and Chen, Jintao and Wang, Xu and Liu, Deyuan and Li, Jungang and Chen, Dechuang and Lin, Ming and Zhou, Jingjiang and Jin, Haopeng and Jia, Qi and Wang, Xiaohang and Wang, Yaole and Zhang, Zhanqiang and Li, Ran and Huang, Zhengkun and Xiong, Shuyue and Wang, Yuji and Dai, Zikun and He, Hui and Luo, Yang and Ning, Mang and Feng, Weiqi and Ye, Chengyang and Lin, Xinyue and Zhao, Min and Zhu, Hongzhou and Tan, Hengkai and Wang, Zeyuan and Xiang, Chendong and Zheng, Kaiwen and Deng, Zhijie and Bao, Fan and Chen, Jianfei and Zhu, Jun},
  journal = {arXiv preprint arXiv:2609.11638},
  year = {2026},
  url = {https://arxiv.org/abs/2609.11638}
}

@article{feng2024tcbench,
  title   = {{TC-Bench: Benchmarking Temporal Compositionality in Text-to-Video and Image-to-Video Generation}},
  author  = {Feng, Weixi and Li, Jiachen and Saxon, Michael and Fu, Tsu-jui and Chen, Wenhu and Wang, William Yang},
  journal = {arXiv preprint arXiv:2406.08656},
  year    = {2024},
  url     = {https://arxiv.org/abs/2406.08656}
}

@article{liu2026iamflow,
  title   = {{Advancing Narrative Long Video Generation via Training-Free Identity-Aware Memory}},
  author  = {Liu, Jinzhuo and Zhang, Jiangning and Jiang, Wencan and Wang, Yabiao and Liang, Dingkang and Xue, Zhucun and Yi, Ran and Liu, Yong},
  journal = {arXiv preprint arXiv:2605.18733},
  year    = {2026},
  url     = {https://arxiv.org/abs/2605.18733}
}

@article{zhang2026mbench,
  title   = {{MBench: A Comprehensive Benchmark on Memory Capability for Video World Models}},
  author  = {Zhang, Shengjun and Zhang, Zhang and Huang, Simin and Tang, Zhenyu and Wang, Hanyang and Dai, Chensheng and Chen, Min and Li, Yifan and Li, Yuxin and Chen, Yingjie and Liu, Hao and Li, Chen and Lyu, Jing and Duan, Yueqi},
  journal = {arXiv preprint arXiv:2606.00793},
  year    = {2026},
  url     = {https://arxiv.org/abs/2606.00793}
}

@article{ying2026wbench,
  title   = {{WBench: A Comprehensive Multi-turn Benchmark for Interactive Video World Model Evaluation}},
  author  = {Ying, Kaining and Hu, Hengrui and Ren, Siyu and Li, Jiamu and Chen, Fengjiao and Wang, Ziwen and Cao, Xuezhi and Cai, Xunliang and Ding, Henghui},
  journal = {arXiv preprint arXiv:2605.25874},
  year    = {2026},
  url     = {https://arxiv.org/abs/2605.25874}
}

@article{xu2026worldmark,
  title   = {{WorldMark: A Unified Benchmark Suite for Interactive Video World Models}},
  author  = {Xu, Xiaojie and Lin, Zhengyuan and He, Kang and Feng, Yukang and Mao, Xiaofeng and Yin, Yuanyang and Ge, Yongtao and Zhang, Kaipeng},
  journal = {arXiv preprint arXiv:2604.21686},
  year    = {2026},
  url     = {https://arxiv.org/abs/2604.21686}
}

@article{li2025egoedit,
  title   = {{EgoEdit: Dataset, Real-Time Streaming Model, and Benchmark for Egocentric Video Editing}},
  author  = {Li, Runjia and Haji-Ali, Moayed and Mirzaei, Ashkan and Wang, Chaoyang and Sahni, Arpit and Skorokhodov, Ivan and Siarohin, Aliaksandr and Jakab, Tomas and Han, Junlin and Tulyakov, Sergey and Torr, Philip and Menapace, Willi},
  journal = {arXiv preprint arXiv:2512.06065},
  year    = {2025},
  url     = {https://arxiv.org/abs/2512.06065}
}

@article{rao2026segue,
  title   = {{No Corners Cut: State-Grounded Transitions for Mid-Stream Prompt Switches in Video Generation}},
  author  = {Rao, Zejing and Ren, Ketong and Liu, Xiaoqiang and Meng, Yiping and Zhang, Guoxin and Tang, Fan},
  journal = {arXiv preprint arXiv:2609.38691},
  year    = {2026},
  url     = {https://arxiv.org/abs/2609.38691}
}

@article{liu2025rollingforcing,
  title   = {{Rolling Forcing: Autoregressive Long Video Diffusion in Real Time}},
  author  = {Liu, Kunhao and Hu, Wenbo and Xu, Jiale and Shan, Ying and Lu, Shijian},
  journal = {arXiv preprint arXiv:2509.25161},
  year    = {2025},
  url     = {https://arxiv.org/abs/2509.25161}
}

@article{ji2025memflow,
  title   = {{MemFlow: Flowing Adaptive Memory for Consistent and Efficient Long Video Narratives}},
  author  = {Ji, Sihui and Chen, Xi and Yang, Shuai and Tao, Xin and Wan, Pengfei and Zhao, Hengshuang},
  journal = {arXiv preprint arXiv:2512.14699},
  year    = {2025},
  url     = {https://arxiv.org/abs/2512.14699}
}

@article{chen2026contextforcing,
  title   = {{Context Forcing: Consistent Autoregressive Video Generation with Long Context}},
  author  = {Chen, Shuo and Wei, Cong and Sun, Sun and Nie, Ping and Zhou, Kai and Zhang, Ge and Yang, Ming-Hsuan and Chen, Wenhu},
  journal = {arXiv preprint arXiv:2602.06028},
  year    = {2026},
  url     = {https://arxiv.org/abs/2602.06028}
}

@article{lu2025rewardforcing,
  title   = {{Reward Forcing: Efficient Streaming Video Generation with Rewarded Distribution Matching Distillation}},
  author  = {Lu, Yunhong and Zeng, Yanhong and Li, Haobo and Ouyang, Hao and Wang, Qiuyu and Cheng, Ka Leong and Zhu, Jiapeng and Cao, Hengyuan and Zhang, Zhipeng and Zhu, Xing and Shen, Yujun and Zhang, Min},
  journal = {arXiv preprint arXiv:2512.04678},
  year    = {2025},
  url     = {https://arxiv.org/abs/2512.04678}
}

@article{cui2026avatarforcing,
  title   = {{AvatarForcing: One-Step Streaming Talking Avatars via Local-Future Sliding-Window Denoising}},
  author  = {Cui, Liyuan and Hu, Wentao and Zhang, Wenyuan and Yang, Zesong and Shi, Fan and Liu, Xiaoqiang},
  journal = {arXiv preprint arXiv:2603.14331},
  year    = {2026},
  url     = {https://arxiv.org/abs/2603.14331}
}

@article{su2026omniforcing,
  title   = {{OmniForcing: Unleashing Real-time Joint Audio-Visual Generation}},
  author  = {Su, Yaofeng and Li, Yuming and Xue, Zeyue and Huang, Jie and Fu, Siming and Li, Haoran and Li, Ying and Qian, Zezhong and Huang, Haoyang and Duan, Nan},
  journal = {arXiv preprint arXiv:2603.11647},
  year    = {2026},
  url     = {https://arxiv.org/abs/2603.11647}
}

@article{liao2026editworld,
  title   = {{Precise Editing and Flexible Referencing for Interactable Worlds}},
  author  = {Liao, Xinyao and Zeng, Xianfang and Liang, Zhu and Fu, Zhoujie and Xu, Qianxun and Liu, Jiachi and Yu, Gang and Lin, Guosheng},
  journal = {arXiv preprint arXiv:2609.34470},
  year    = {2026},
  url     = {https://arxiv.org/abs/2609.34470}
}

@article{zhao2026causalforcingpp,
  title   = {{Causal Forcing++: Scalable Few-Step Autoregressive Diffusion Distillation for Real-Time Interactive Video Generation}},
  author  = {Zhao, Min and Zhu, Hongzhou and Zheng, Kaiwen and Zhou, Zihan and Yan, Bokai and Li, Xinyuan and Yang, Xiao and Li, Chongxuan and Zhu, Jun},
  journal = {arXiv preprint arXiv:2605.15141},
  year    = {2026},
  url     = {https://arxiv.org/abs/2605.15141}
}

@article{mao2025yume15,
  title   = {{Yume-1.5: A Text-Controlled Interactive World Generation Model}},
  author  = {Mao, Xiaofeng and Li, Zhen and Li, Chuanhao and Xu, Xiaojie and Ying, Kaining and He, Tong and Pang, Jiangmiao and Qiao, Yu and Zhang, Kaipeng},
  journal = {arXiv preprint arXiv:2512.22096},
  year    = {2025},
  url     = {https://arxiv.org/abs/2512.22096}
}

@article{zhen2026soulxliveact,
  title   = {{SoulX-LiveAct: Towards Hour-Scale Real-Time Human Animation with Neighbor Forcing and ConvKV Memory}},
  author  = {Zhen, Dingcheng and Zheng, Xu and Zhang, Ruixin and Jiang, Zhiqi and Yan, Yichao and Tao, Ming and Yin, Shunshun},
  journal = {arXiv preprint arXiv:2603.11746},
  year    = {2026},
  url     = {https://arxiv.org/abs/2603.11746}
}

@article{zhang2025musetalk,
  title   = {{MuseTalk: Real-Time High-Fidelity Video Dubbing via Spatio-Temporal Sampling}},
  author  = {Zhang, Yue and Zhong, Zhizhou and Liu, Minhao and Chen, Zhaokang and Wu, Bin and Zeng, Yubin and Zhan, Chao and He, Yingjie and Huang, Junxin and Zhou, Wenjiang},
  journal = {arXiv preprint arXiv:2410.10122v3},
  year    = {2025},
  url     = {https://arxiv.org/abs/2410.10122v3}
}

@inproceedings{zhang2025spargeattention,
  title = {{SpargeAttention: Accurate and Training-free Sparse Attention Accelerating Any Model Inference}},
  author = {Zhang, Jintao and Xiang, Chendong and Huang, Haofeng and Wei, Jia and Xi, Haocheng and Zhu, Jun and Chen, Jianfei},
  booktitle = {Proceedings of the 42nd International Conference on Machine Learning},
  year = {2025},
  url = {https://proceedings.mlr.press/v267/zhang25ch.html},
  pages = {76397--76413},
  volume = {267},
  series = {Proceedings of Machine Learning Research},
  publisher = {PMLR}
}

@inproceedings{xi2025sparsevideogen,
  title = {{Sparse Video-Gen: Accelerating Video Diffusion Transformers with Spatial-Temporal Sparsity}},
  author = {Xi, Haocheng and Yang, Shuo and Zhao, Yilong and Xu, Chenfeng and Li, Muyang and Li, Xiuyu and Lin, Yujun and Cai, Han and Zhang, Jintao and Li, Dacheng and Chen, Jianfei and Stoica, Ion and Keutzer, Kurt and Han, Song},
  booktitle = {Proceedings of the 42nd International Conference on Machine Learning},
  year = {2025},
  url = {https://proceedings.mlr.press/v267/xi25c.html},
  pages = {68208--68224},
  volume = {267},
  series = {Proceedings of Machine Learning Research},
  publisher = {PMLR}
}

@inproceedings{zhang2025sageattention,
  author = {Zhang, Jintao and Wei, Jia and Zhang, Pengle and Zhu, Jun and Chen, Jianfei},
  booktitle = {International Conference on Learning Representations},
  editor = {Y. Yue and A. Garg and N. Peng and F. Sha and R. Yu},
  pages = {71566--71585},
  title = {{SageAttention: Accurate 8-Bit Attention for Plug-and-play Inference Acceleration}},
  url = {https://proceedings.iclr.cc/paper_files/paper/2025/file/b286c344d38e10d2466c0514b78e2f36-Paper-Conference.pdf},
  volume = {2025},
  year = {2025}
}

@inproceedings{zhang2025sageattention2,
  title = {{SageAttention2: Efficient Attention with Thorough Outlier Smoothing and Per-thread INT4 Quantization}},
  author = {Zhang, Jintao and Huang, Haofeng and Zhang, Pengle and Wei, Jia and Zhu, Jun and Chen, Jianfei},
  booktitle = {Proceedings of the 42nd International Conference on Machine Learning},
  year = {2025},
  url = {https://proceedings.mlr.press/v267/zhang25ae.html},
  pages = {75097--75119},
  volume = {267},
  series = {Proceedings of Machine Learning Research},
  publisher = {PMLR}
}

@inproceedings{yang2025sparsevideogen2,
  author = {Yang, Shuo and Xi, Haocheng and Zhao, Yilong and Li, Muyang and Zhang, Jintao and Cai, Han and Lin, Yujun and Li, Xiuyu and Xu, Chenfeng and Peng, Kelly and Chen, Jianfei and Han, Song and Keutzer, Kurt and Stoica, Ion},
  booktitle = {Advances in Neural Information Processing Systems},
  doi = {10.52202/085713-3243},
  editor = {D. Belgrave and C. Zhang and H. Lin and R. Pascanu and P. Koniusz and M. Ghassemi and N. Chen},
  pages = {96965--96991},
  publisher = {Curran Associates, Inc.},
  title = {{Sparse VideoGen2: Accelerate Video Generation with  Sparse Attention via Semantic-Aware Permutation}},
  url = {https://proceedings.neurips.cc/paper_files/paper/2025/file/8bd5148caced2d73cea7b6961a874a49-Paper-Conference.pdf},
  volume = {38, Main Conference},
  year = {2025}
}

@inproceedings{zhang2025sageattention3,
  author = {Zhang, Jintao and Wei, Jia and Wang, Haoxu and Zhang, Pengle and Xu, Xiaoming and Huang, Haofeng and Jiang, Kai and Chen, Jianfei and Zhu, Jun},
  booktitle = {Advances in Neural Information Processing Systems},
  doi = {10.52202/085713-1799},
  editor = {D. Belgrave and C. Zhang and H. Lin and R. Pascanu and P. Koniusz and M. Ghassemi and N. Chen},
  pages = {53901--53931},
  publisher = {Curran Associates, Inc.},
  title = {{SageAttention3: Microscaling FP4 Attention for Inference and An Exploration of 8-Bit Training}},
  url = {https://proceedings.neurips.cc/paper_files/paper/2025/file/4db397e0f760cc573c681e81a01a3dba-Paper-Conference.pdf},
  volume = {38, Main Conference},
  year = {2025}
}

@inproceedings{zheng2026rcm,
  author = {Zheng, Kaiwen and Wang, Yuji and Ma, Qianli and Chen, Huayu and Zhang, Jintao and Balaji, Yogesh and Chen, Jianfei and Liu, Ming-Yu and Zhu, Jun and Zhang, Qinsheng},
  booktitle = {International Conference on Learning Representations},
  editor = {C. Vondrick and B. Hariharan and C. Raffel and L. Pinto and D. Yang and A. Faust},
  pages = {2582--2603},
  title = {{Large Scale Diffusion Distillation via Score-Regularized Continuous-Time Consistency}},
  url = {https://proceedings.iclr.cc/paper_files/paper/2026/file/0534abc9e6db91683d82186ef0d68202-Paper-Conference.pdf},
  volume = {2026},
  year = {2026}
}

@inproceedings{zhang2026sla,
  author = {Zhang, Jintao and Wang, Haoxu and Jiang, Kai and Yang, Shuo and Zheng, Kaiwen and Xi, Haocheng and Wang, Ziteng and Zhu, Hongzhou and Zhao, Min and Stoica, Ion and Gonzalez, Joseph E and Chen, Jianfei and Zhu, Jun},
  booktitle = {International Conference on Learning Representations},
  editor = {C. Vondrick and B. Hariharan and C. Raffel and L. Pinto and D. Yang and A. Faust},
  pages = {138288--138305},
  title = {{SLA: Beyond Sparsity in Diffusion Transformers via Fine-Tunable Sparse--Linear Attention}},
  url = {https://proceedings.iclr.cc/paper_files/paper/2026/file/df77cca99e2785eafb5b2416d688ae24-Paper-Conference.pdf},
  volume = {2026},
  year = {2026}
}

@article{zhang2025turbodiffusion,
  title = {{TurboDiffusion: Accelerating Video Diffusion Models by 100-200 Times}},
  author = {Zhang, Jintao and Zheng, Kaiwen and Jiang, Kai and Wang, Haoxu and Stoica, Ion and Gonzalez, Joseph E and Chen, Jianfei and Zhu, Jun},
  journal = {arXiv preprint arXiv:2512.16093},
  year = {2025},
  url = {https://arxiv.org/abs/2512.16093}
}

@article{zhang2025sageattention2pp,
  title = {{SageAttention2++: A More Efficient Implementation of SageAttention2}},
  author = {Zhang, Jintao and Xu, Xiaoming and Wei, Jia and Huang, Haofeng and Zhang, Pengle and Xiang, Chendong and Zhu, Jun and Chen, Jianfei},
  journal = {arXiv preprint arXiv:2505.21136},
  year = {2025},
  url = {https://arxiv.org/abs/2505.21136}
}

@article{zheng2026causalrcm,
  title = {{Causal-rCM: A Unified Teacher-Forcing and Self-Forcing Open Recipe for Autoregressive Diffusion Distillation in Streaming Video Generation and Interactive World Models}},
  author = {Zheng, Kaiwen and He, Guande and Zhao, Min and Zhang, Jintao and Chen, Huayu and Chen, Jianfei and Lin, Chen-Hsuan and Liu, Ming-Yu and Zhu, Jun and Ma, Qianli},
  journal = {arXiv preprint arXiv:2606.25473},
  year = {2026},
  url = {https://arxiv.org/abs/2606.25473}
}

@article{xi2026quantvideogen,
  title = {{Quant VideoGen: Auto-Regressive Long Video Generation via 2-Bit KV-Cache Quantization}},
  author = {Xi, Haocheng and Yang, Shuo and Zhao, Yilong and Li, Muyang and Cai, Han and Li, Xingyang and Lin, Yujun and Zhang, Zhuoyang and Zhang, Jintao and Li, Xiuyu and Xu, Zhiying and Wu, Jun and Xu, Chenfeng and Stoica, Ion and Han, Song and Keutzer, Kurt},
  journal = {arXiv preprint arXiv:2602.02958},
  year = {2026},
  url = {https://arxiv.org/abs/2602.02958}
}

@article{zhang2026sla2,
  title = {{SLA2: Sparse-Linear Attention with Learnable Routing and QAT}},
  author = {Zhang, Jintao and Wang, Haoxu and Jiang, Kai and Zheng, Kaiwen and Jiang, Youhe and Stoica, Ion and Chen, Jianfei and Zhu, Jun and Gonzalez, Joseph E},
  year = {2026},
  url = {https://arxiv.org/abs/2602.12675},
  journal = {arXiv preprint arXiv:2602.12675}
}

@inproceedings{zhao2026ultravico,
  author = {Zhao, Min and Zhu, Hongzhou and Wang, Yingze and Yan, Bokai and Zhang, Jintao and He, Guande and Yang, Ling and Li, Chongxuan and Zhu, Jun},
  booktitle = {International Conference on Learning Representations},
  editor = {C. Vondrick and B. Hariharan and C. Raffel and L. Pinto and D. Yang and A. Faust},
  pages = {122283--122309},
  title = {{UltraViCo: Breaking Extrapolation Limits in Video Diffusion Transformers}},
  url = {https://proceedings.iclr.cc/paper_files/paper/2026/file/c6e3856954d23bec921f2d13d8c0e0e7-Paper-Conference.pdf},
  volume = {2026},
  year = {2026}
}

@article{zhang2026spargeattention2,
  title = {{SpargeAttention2: Trainable Sparse Attention via Hybrid Top-k+Top-p Masking and Distillation Fine-Tuning}},
  author = {Zhang, Jintao and Jiang, Kai and Xiang, Chendong and Feng, Weiqi and Hu, Yuezhou and Xi, Haocheng and Chen, Jianfei and Zhu, Jun},
  year = {2026},
  url = {https://arxiv.org/abs/2602.13515},
  journal = {arXiv preprint arXiv:2602.13515}
}

@article{xiang2026viewrope,
  title = {{Geometry-Aware Rotary Position Embedding for Consistent Video World Model}},
  author = {Xiang, Chendong and Liu, Jiajun and Zhang, Jintao and Yang, Xiao and Fang, Zhengwei and Wang, Shizun and Wang, Zijun and Zou, Yingtian and Su, Hang and Zhu, Jun},
  journal = {arXiv preprint arXiv:2602.07854},
  year = {2026},
  url = {https://arxiv.org/abs/2602.07854}
}

@misc{zhang2025efficientattention,
  title = {{Efficient Attention Methods: Hardware-efficient, Sparse, Compact, and Linear Attention}},
  author = {Zhang, Jintao and Su, Rundong and Liu, Chunyu and Wei, Jia and Wang, Ziteng and Wang, Haoxu and Zhang, Pengle and Jiang, Huiqiang and Huang, Haofeng and Xiang, Chendong and Xi, Haocheng and Yang, Shuo and Li, Xingyang and Zhao, Yilong and Hu, Yuezhou and Fu, Tianyu and Zhao, Tianchen and Zhang, Yicheng and Cao, Boqun and Jiang, Youhe and Chen, Chang and Jiang, Kai and Chen, Huayu and Zhao, Min and Xu, Xiaoming and Wu, Yi and Bao, Fan and Stoica, Ion and Gonzalez, Joseph E. and Chen, Jianfei and Zhu, Jun},
  year = {2025},
  howpublished = {Technical report},
  url = {https://attention-survey.github.io/files/Attention_Survey.pdf}
}

@article{jiang2026turboserve,
  title = {{TurboServe: Serving Streaming Video Generation Efficiently and Economically}},
  author = {Jiang, Youhe and Wang, Haoxu and Bao, Haotong and Jiang, Kai and Chen, Jianfei and Zhu, Jun and Fu, Fangcheng and Zhang, Jintao},
  journal = {arXiv preprint arXiv:2606.19271},
  year = {2026},
  url = {https://arxiv.org/abs/2606.19271}
}

@article{hu2026sdvg,
  title = {{Speculative Decoding for Autoregressive Video Generation}},
  author = {Hu, Yuezhou and Zhang, Jintao},
  journal = {arXiv preprint arXiv:2604.17397},
  year = {2026},
  url = {https://arxiv.org/abs/2604.17397}
}

@InProceedings{zhou2026svgear,
  author = {Zhou, Xuanyi
and Mang, Qiuyang
and Yang, Shuo
and Xi, Haocheng
and Zhang, Jintao
and Mao, Huanzhi
and Gonzalez, Joseph E.
and Keutzer, Kurt
and Stoica, Ion
and Cheung, Alvin},
  editor = {Favaro, Paolo
and Kukelova, Zuzana
and Maki, Atsuto
and Rohrbach, Anna
and Schindler, Konrad
and Tombari, Federico},
  title = {{SVG-EAR: Parameter-Free Linear Compensation for Sparse Video Generation via Error-aware Routing}},
  booktitle = {Computer Vision -- ECCV 2026},
  year = {2026},
  publisher = {Springer Nature Switzerland},
  address = {Cham},
  pages = {167--185},
  isbn = {978-3-032-37359-5},
  doi = {10.1007/978-3-032-37359-5_10},
  url = {https://link.springer.com/chapter/10.1007/978-3-032-37359-5_10}
}

@article{su2026sixbitdiffusion,
  title = {{6Bit-Diffusion: Inference-Time Mixed-Precision Quantization for Video Diffusion Models}},
  author = {Su, Rundong and Zhang, Jintao and Yuan, Zhihang and Duanmu, Haojie and Chen, Jianfei and Zhu, Jun},
  journal = {arXiv preprint arXiv:2603.18742},
  year = {2026},
  url = {https://arxiv.org/abs/2603.18742}
}

@article{yang2026turbot2va,
  title = {{TurboT2VA: Fast Large-Scale Text-to-Video-Audio Generation via Score-Regularized Consistency Distillation}},
  author = {Yang, Xiaoda and Liu, Yuxiang and Zheng, Kaiwen and Liu, Yuan and Lai, Yibo and Ji, Shengpeng and Jiang, Kai and Chen, Jianfei and Yang, Shan and Liang, Sen and Hu, Xiaobin and Yan, Shuicheng and Zhang, Jintao and Zhu, Jun and Zhao, Zhou},
  journal = {arXiv preprint arXiv:2608.24674},
  year = {2026},
  url = {https://arxiv.org/abs/2608.24674}
}

@misc{zhuGeneralWorldModels,
  title = {General World Models from First Principles},
  author = {Zhu, Jun and Tan, Hengkai and Zhang, Jintao and Zhao, Min and Bao, Fan and Zhang, Bo},
  howpublished = {Available at SSRN 7412285},
  url = {https://papers.ssrn.com/sol3/papers.cfm?abstract_id=7412285},
  year = {2026},
  doi = {10.2139/ssrn.7412285}
}

@article{zhang2025int8fallback,
  title = {{Accurate INT8 Training Through Dynamic Block-Level Fallback}},
  author = {Zhang, Pengle and Wei, Jia and Zhang, Jintao and Zhu, Jun and Chen, Jianfei},
  journal = {arXiv preprint arXiv:2503.08040},
  year = {2025},
  url = {https://arxiv.org/abs/2503.08040}
}

@article{wang2026kernelbenchx,
  title = {{KernelBenchX: A Comprehensive Benchmark for Evaluating LLM-Generated GPU Kernels}},
  author = {Wang, Han and Zhang, Jintao and Jiang, Kai and Wang, Haoxu and Chen, Jianfei and Zhu, Jun},
  journal = {arXiv preprint arXiv:2605.04956},
  year = {2026},
  url = {https://arxiv.org/abs/2605.04956}
}

@article{zhang2026sagebwd,
  title = {{SageBwd: A Trainable Low-bit Attention}},
  author = {Zhang, Jintao and Chen, Marco and Wang, Haoxu and Jiang, Kai and Stoica, Ion and Gonzalez, Joseph E. and Chen, Jianfei and Zhu, Jun},
  journal = {arXiv preprint arXiv:2603.02170},
  year = {2026},
  url = {https://arxiv.org/abs/2603.02170}
}

@article{yang2026flashkmeans,
  title = {{Flash-KMeans: Fast and Memory-Efficient Exact K-Means}},
  author = {Yang, Shuo and Xi, Haocheng and Zhao, Yilong and Li, Muyang and Fan, Xiaoze and Zhang, Jintao and Cai, Han and Lin, Yujun and Li, Xiuyu and Keutzer, Kurt and Han, Song and Xu, Chenfeng and Stoica, Ion},
  journal = {arXiv preprint arXiv:2603.09229},
  year = {2026},
  url = {https://arxiv.org/abs/2603.09229}
}

@article{mazumdar2026videofdb,
  title   = {{VideoFDB: Evaluating Full-Duplex Vision-Speech Capabilities in Conversational Agents}},
  author  = {Mazumdar, Amrita and Park, Seonwook and Roy, Rajarshi and Srihari, Nikhil and Wang, Shengze and Zhou, Yuhao and Wang, Julia and Nagano, Koki and De Mello, Shalini},
  journal = {arXiv preprint arXiv:2605.30256},
  year    = {2026},
  url     = {https://arxiv.org/abs/2605.30256v2}
}

@inproceedings{zhang2025sagerag,
  title = {{SAGE: A Framework of Precise Retrieval for RAG}},
  author = {Zhang, Jintao and Li, Guoliang and Su, Jinyang},
  booktitle = {2025 IEEE 41st International Conference on Data Engineering (ICDE)},
  year = {2025},
  url = {https://ieeexplore.ieee.org/abstract/document/11112897/}
}

@article{jiang2025cascadia,
  title = {{Cascadia: An efficient cascade serving system for large language models}},
  author = {Jiang, Youhe and Fu, Fangcheng and Zhao, Wanru and Rabanser, Stephan and Zhang, Jintao and Lane, Nicholas D and Yuan, Binhang},
  journal = {arXiv preprint arXiv:2506.04203},
  year = {2025},
  url = {https://arxiv.org/abs/2506.04203}
}

@article{hu2026residualcontext,
  title = {{Residual Context Diffusion Language Models}},
  author = {Hu, Yuezhou and Singh, Harman and Maheswaran, Monishwaran and Xi, Haocheng and Hooper, Coleman and Zhang, Jintao and Tomar, Aditya and Mahoney, Michael W and Min, Sewon and Farajtabar, Mehrdad and Keutzer, Kurt and Gholami, Amir and Xu, Chenfeng},
  journal = {arXiv preprint arXiv:2601.22954},
  year = {2026},
  url = {https://arxiv.org/abs/2601.22954}
}

@article{zhao2025ultraimage,
  title = {{Ultraimage: Rethinking resolution extrapolation in image diffusion transformers}},
  author = {Zhao, Min and Yan, Bokai and Yang, Xue and Zhu, Hongzhou and Zhang, Jintao and Liu, Shilong and Li, Chongxuan and Zhu, Jun},
  journal = {arXiv preprint arXiv:2512.04504},
  year = {2025},
  url = {https://arxiv.org/abs/2512.04504}
}

@article{hu2025postquantization,
  title = {{Identifying sensitive weights via post-quantization integral}},
  author = {Hu, Yuezhou and Huang, Weiyu and Liang, Zichen and Chen, Chang and Zhang, Jintao and Zhu, Jun and Chen, Jianfei},
  journal = {arXiv preprint arXiv:2503.01901},
  year = {2025},
  url = {https://arxiv.org/abs/2503.01901}
}

@article{sun2026hybridstitch,
  title = {{HybridStitch: Pixel and Timestep Level Model Stitching for Diffusion Acceleration}},
  author = {Sun, Desen and Hon, Jason and Zhang, Jintao and Liu, Sihang},
  journal = {arXiv preprint arXiv:2603.07815},
  year = {2026},
  url = {https://arxiv.org/abs/2603.07815}
}

@inproceedings{jiang2026hexgen3,
  title = {{HexGen-3: A Fully Disaggregated LLM Serving Framework with Fine-Grained Heterogeneous Resource Autoscaling}},
  author = {Jiang, Youhe and Li, Wenshuang and Peng, You and Zhang, Jintao and Yan, Ran and Chen, Jianfei and Han, Xu and Fu, Fangcheng and Yuan, Binhang},
  booktitle = {Forty-third International Conference on Machine Learning},
  year = {2026},
  url = {https://openreview.net/forum?id=fVG57FKBNp}
}

\section{Contributions and Acknowledgments}
\label{sec:contributions_acknowledgments}

\subsection*{StreamEval Team}
\begingroup
\setlength{\parskip}{0pt}
{\sffamily\bfseries\color{reportblue}
\fontsize{9.36}{16}\selectfont
\raggedright
\noindent
\mbox{Jungang Li},\hspace{0.75em}%
\mbox{Ziqi Huang},\hspace{0.75em}%
\mbox{Tian Ye},\hspace{0.75em}%
\mbox{Haocheng Xi},\hspace{0.75em}%
\mbox{Yuezhou Hu},\hspace{0.75em}%
\mbox{Kai Jiang},\hspace{0.75em}%
\mbox{Zhifei Li},\hspace{0.75em}%
\mbox{Yichuan Wang},\hspace{0.75em}%
\mbox{Shuo Yang},\hspace{0.75em}%
\mbox{Zhiqiu Lin},\hspace{0.75em}%
\mbox{Zian Zheng},\hspace{0.75em}%
\mbox{Alberto Taiuti},\hspace{0.75em}%
\mbox{Jintao Zhang},\hspace{0.75em}%
\mbox{Jianfei Chen},\hspace{0.75em}%
\mbox{Jun Zhu}
\par}

\vspace{4pt}
\noindent
{\normalfont\small\itshape\color{black!65}
Author order does not reflect relative contributions.
\par}
\endgroup

\subsection*{Acknowledgments}
We thank ShengShu Technology for providing computational resources.
We gratefully acknowledge the support of our partner institutions:
Tsinghua University, NVIDIA, Carnegie Mellon University,
the University of California, Berkeley, Princeton University,
the National University of Singapore, and Reactor.
We also thank all collaborators for their help and contributions.

\end{document}